\documentclass[lettersize,journal]{IEEEtran}
\usepackage{amsmath,amsfonts,amssymb}
\usepackage{booktabs}
\usepackage{pifont}
\usepackage{algorithmic}
\usepackage{algorithm}
\usepackage{multirow,array}
\usepackage[caption=false,font=normalsize,labelfont=sf,textfont=sf]{subfig}
\usepackage{textcomp}
\usepackage{stfloats}
\usepackage{makecell}
\usepackage{url}
\usepackage{verbatim}
\usepackage{graphicx}
\usepackage[normalem]{ulem}
\usepackage{cite}
\usepackage{xcolor}
\usepackage{soul}
\usepackage{hyperref}
\usepackage{twemojis}
\usepackage{tikz}
\usepackage{eso-pic}
\newcommand{\copyrightnotice}{
  \AddToShipoutPictureFG*{%
    \AtPageUpperLeft{%
      \raisebox{-35pt}[0pt][0pt]{%
        \hspace{60pt}%
        \begin{minipage}{0.95\textwidth}
          \footnotesize \textcolor{gray}{%
            \textcopyright~2024 IEEE. Personal use of this material is permitted. 
            Permission from IEEE must be obtained for all other uses, in any current or future media, including reprinting/republishing this material for advertising or promotional purposes, creating new collective works, for resale or redistribution to servers or lists, or reuse of any copyrighted component of this work in other works. This article has been accepted for publication in \textit{IEEE Transactions on Image Processing}, vol.~33, pp.~6865 -- 6880, 2024. DOI: \url{10.1109/TIP.2024.3512369}
          }
        \end{minipage}
      }
    }
  }
}

\begin{document}

\title{Advancing Video Anomaly Detection: A Bi-Directional Hybrid Framework for Enhanced Single- and Multi-Task Approaches}

\author{Guodong Shen, Yuqi Ouyang, Junru Lu, Yixuan Yang, and Victor Sanchez, \IEEEmembership{Member,~IEEE}
\thanks{Guodong Shen, Yuqi Ouyang, Junru Lu, Yixuan Yang, and Victor Sanchez are with the Computer Science Department at the University of Warwick, Coventry, United Kingdom (e-mail: guodong.shen@warwick.ac.uk; yuqi.ouyang@warwick.ac.uk; lj1230@nyu.edu; arnoldyang97@gmail.com; v.f.sanchez-silva@warwick.ac.uk). Yuqi Ouyang was a PhD student at the University of Warwick when this work was completed and is now with the College of Computer Science, Sichuan University, Chengdu, China (e-mail: yuqi.ouyang@scu.edu.cn).}
}



\copyrightnotice
\maketitle

\begin{abstract}

Despite the prevailing transition from single-task to multi-task approaches in video anomaly detection, 
 we observe that many adopt sub-optimal frameworks for individual proxy tasks. Motivated by this, we contend that optimizing single-task frameworks can advance both single- and multi-task approaches. Accordingly, we leverage middle-frame prediction as the primary proxy task, and introduce an effective hybrid framework designed to generate accurate predictions for normal frames and flawed predictions for abnormal frames.  This hybrid framework is built upon a bi-directional structure that seamlessly integrates both vision transformers and ConvLSTMs. Specifically, we utilize this bi-directional structure to fully analyze the temporal dimension by predicting frames in both forward and backward directions, significantly boosting the detection stability. Given the transformer's capacity to model long-range contextual dependencies, we develop a convolutional temporal transformer that efficiently associates feature maps from all context frames to generate attention-based predictions for target frames. Furthermore, we devise a layer-interactive ConvLSTM bridge that facilitates the smooth flow of low-level features across layers and time-steps, thereby strengthening predictions with fine details. Anomalies are eventually identified by scrutinizing the discrepancies between target frames and their corresponding predictions. Several experiments conducted on public benchmarks affirm the efficacy of our hybrid framework, whether used as a standalone single-task approach or integrated as a branch in a multi-task approach.  These experiments also underscore the advantages of merging vision transformers and ConvLSTMs for video anomaly detection. The implementation of our hybrid framework is available at \url{https://github.com/SHENGUODONG19951126/ConvTTrans-ConvLSTM}.



\end{abstract}


\section{Introduction\label{sec:intro}}
\IEEEPARstart{V}{ideo} anomaly detection (VAD) has experienced significant progress in recent few years, holding the prospect for bolstering intelligent surveillance systems over the long term.  However, the task remains challenging due to the rarity, ambiguity, and innumerable nature of anomalous instances \cite{zaheer2020old,park2020learning,feng2021mist}. Consequently, weakly-supervised \cite{zhong2019graph,wan2020weakly,sultani2018real,zhang2019temporal,feng2021mist,zaheer2020claws} and self-supervised \cite{zaheer2022generative,pang2020self,Lin_Chen_Li_Yu_2022, tudor2017unmasking, wang2018detecting} solutions, which rely on anomalous data during training, often struggle to handle real-world scenarios where countless ``unseen'' anomalies may occur \cite{doshi2020continual,feng2021convolutional,liu2019margin,zaheer2022generative}. As Zhu et al. \cite{zhu2022towards} validate, these solutions even underperform on VAD benchmarks when exposed to a limited number of training anomalies. 
In response to this deviation between benchmarks and real-world scenarios, recent research has witnessed a resurgence in One-class Classifiers (OCCs) \cite{deng2023bi,Szymanowicz2021XMANEM,leroux2022multi,szymanowicz2022discrete,Park2022FastAnoFA,lv2021learning,cai2021appearance}, which are trained solely on normal data and identify divergent patterns as anomalies during testing.



Based on the number of proxy tasks \cite{georgescu2021anomaly} used to learn abnormalities, existing OCCs can be broadly categorized as single-task or multi-task approaches. 
Single-task approaches rely on an individual proxy task to estimate the target VAD task.  They are primarily driven by reconstruction and prediction models. Reconstruction models typically use U-nets and Convolutional Auto-Encoders to embed frames into compact feature maps through frame reconstruction. They infer anomalies by measuring either reconstruction errors \cite{liu2018future,tang2020integrating,park2020learning,gong2019memorizing,zaheer2020old,nguyen2019anomaly} or distribution discrepancies in the latent space \cite{ouyang2021video,Xu2015LearningDR,ionescu2019object}. However, since this proxy task requires the output to match the input, models with strong generalization capabilities sometimes inadvertently produce accurate reconstructions for anomalies\cite{liu2018future}. To alleviate this issue, prediction models \cite{shen2022video,liu2021hybrid,georgescu2021anomaly,lv2021learning,cai2021appearance,lai2020video,feng2021convolutional,yuan2021transanomaly,tang2020integrating,liu2018future} interpret anomalies as unpredictable events and explore intrinsic spatio-temporal dynamics through frame prediction. The indirectness of the prediction process restrains  models from generalizing too well to anomalies \cite{liu2018future}, thus enlarging the disparity between normal and abnormal events. 

On the other hand, multi-task approaches rely on a set of proxy tasks to compensate for the misalignment between each individual proxy task and the target VAD task \cite{georgescu2021anomaly}. Initially conceived as multi-branch models, they have recently gained prominence following the work by Georgescu et al.  \cite{georgescu2021anomaly}. In multi-task approaches, each branch corresponds to an individual proxy task.  
These proxy tasks frequently encompass the aforementioned frame reconstruction and prediction \cite{lai2020video,luo2017remembering}, as well as tasks related to higher-level descriptors such as optical flow \cite{nguyen2019anomaly,cai2021appearance,georgescu2021background}, clustering \cite{chang2020clustering}, conditional probability density \cite{abati2019latent}, gradient \cite{yu2020cloze, ionescu2019object}, human skeleton \cite{doshi2023towards}, and difference maps \cite{chang2020clustering}.  Notably, Georgescu et al. \cite{georgescu2021anomaly} further enrich the scope of proxy tasks by formulating tasks related to the temporal dimension, motion continuity, and model distillation. By sharing minor network modules across tasks, multi-task approaches are expected to extract well-generalized prior knowledge \cite{Caruana1997MultitaskL}, resulting in improved performance for {\it each} proxy task or branch compared to their counterparts in single-task approaches. However,  many multi-task approaches fail to meet this expectation despite their impressive overall performance in anomaly inference \cite{georgescu2021anomaly,georgescu2021background,ionescu2019object}. For instance, when considering only the middle-frame prediction, Georgescu et al.'s multi-task branch \cite{georgescu2021anomaly} exhibits inferior performance than that of Deng et al.'s single-task approach \cite{deng2023bi} on the UCSD Ped2 and CUHK Avenue datasets. This observation suggests that existing multi-task approaches may adopt sub-optimal frameworks for individual proxy tasks, leaving potential space for improvement.


 Based on the previous discussions, we argue that refining single-task frameworks could benefit both single- and multi-task VAD approaches. 
 An ideal framework should detect anomalies accurately as a single-task approach while serving as a strong-performing branch in multi-task approaches. To this end, we focus on middle-frame prediction as the proxy task by following \cite{deng2023bi,georgescu2021anomaly} and propose a hybrid framework that incorporates recent advancements in vision transformers and ConvLSTMs. Instead of relying on existing paradigms,  we have meticulously crafted and tailored all modules for the VAD purpose. First, the proposed framework adopts a bi-directional structure consisting of an encoding pipeline and twin decoding pipelines. This structure enables our framework to predict middle frames in both the forward and backward temporal directions, differentiating itself from other approaches that advocate uni-directional prediction\cite{liu2021hybrid,georgescu2021anomaly,cai2021appearance,lai2020video}. As a result, the framework showcases increased robustness along the temporal dimension, where motion irregularities commonly manifest, thus reinforcing the VAD process against noise and fluctuations. Second, we introduce a novel convolutional temporal transformer as the framework's backbone.  Unlike the previous ViT \cite{dosovitskiy2020image} and its subsequent variations \cite{arnab2021vivit,wang2022bevt}, our transformer focuses on capturing long-range {\it inter-frame} dependencies and generating frame-wise predictions. To this end, it incorporates a unique temporal self-attention mechanism, which processes the global spatial information of whole frames without the need for patch division \cite{dosovitskiy2020image,arnab2021vivit,wang2022bevt
 }. In practice, the transformer encoder  extracts contextual knowledge from context clips in parallel, while the decoder exploits this knowledge to generate a prediction sequence for the target clip.  
Third, we introduce a layer-interactive ConvLSTM bridge that spans each decoding pipeline to efficiently transmit low-level features across layers and time-steps. This mechanism prevents the loss of crucial low-level features as they undergo further compression by the transformer, enabling the framework to produce finer-grained predictions to better distinguish irregularities. Notably, our framework stands out as the first to integrate two prediction mechanisms, namely vision transformers and ConvLSTMs, with the purpose of harnessing and merging their respective strengths. While our convolutional transformer specializes in modeling long-range global dependencies, our ConvLSTM bridge excels at capturing short-range local details. After a standard one-stage training process through a combined objective function, our framework consistently attains compelling outcomes on public VAD benchmarks, including UCSD Ped2 \cite{mahadevan2010anomaly}, CUHK Avenue \cite{lu2013abnormal}, ShanghaiTech \cite{luo2017revisit}, and Street Scene \cite{ramachandra2020street}.  Specifically, our framework's efficacy is observed both when employed independently as a single-task approach and jointly as a branch in a multi-task approach. The contributions of this work are then manifold:

1) We utilize middle-frame prediction as the proxy task and propose a hybrid framework to advance both single- and multi-task VAD approaches.

2) We incorporate a bi-directional structure with two decoding pipelines. This design enables our framework to predict in opposite temporal directions, reinforcing its resilience against motion irregularities and improving inference stability.

3) We introduce a convolutional temporal transformer to model long-range inter-frame dependencies. Its unique temporal self-attention mechanism efficiently processes entire frames without patch division, thus preserving global spatial information for frame-wise prediction. 

4) We introduce a layer-interactive ConvLSTM bridge to effectively propagate low-level features across layers and time-steps, thereby refining predictions with granular details. 

5) We integrate vision transformers and ConvLSTMs within a unified framework to exploit their respective strengths for improved VAD performance.


6) We conduct comprehensive experiments on standard benchmarks, arguing the framework's competitiveness against other state-of-the-art approaches, either implemented as a single-task approach or as a branch in a multi-task approach. 

The rest of the paper is organized as follows. Section \ref{sec:relawork} critically examines related work. Section \ref{sec:profra} elaborates on the architecture of our proposed framework. Section \ref{sec:experi} presents a series of experiments conducted for framework evaluation, detailing the experimental setup, results, and their interpretation. The paper concludes in Section \ref{sec:conclusion}.

\section{Related Work}
\label{sec:relawork}

 \subsection{Vision Transformer} Transformers have revolutionized and dominated modern prediction tasks across various domains, including engineering system monitoring\cite{cao2024advanced}, cyberattack detection\cite{xu2021hybrid}, and even the groundbreaking ChatGPT\cite{openai2023chatgpt}, owing to their capability to model long-range dependencies. In computer vision, vision transformers, evolved from the original transformer architecture \cite{vaswani2017attention}, have also gained widespread use, particularly in image classification and recognition tasks\cite{dosovitskiy2020image, arnab2021vivit, wang2022bevt, liu2021swin, fan2021multiscale, meng2022adavit, parmar2018image}. However, handling entire images as single entities remains challenging due to  the high dimensionality of imagery data.  To overcome this challenge, existing vision transformers like ViT\cite{dosovitskiy2020image},  ViViT \cite{arnab2021vivit}, and BEVT \cite{wang2022bevt}, divide images into  sequences of patches and extract regional features from each patch.  
While this patch-based strategy proves effective for classification and recognition tasks that merely involve feature extraction, it poses difficulties for prediction-based VAD, which requires rebuilding cross-patch relationships to predict an entire frame \cite{lee2022multi}.

As a result, despite their irreversible momentum and vast potential, vision transformers are {\it seldom} investigated in the field of prediction-based VAD. Among the rare VAD works inspired by transformers, CT-D2GAN \cite{feng2021convolutional} inherits positional encoding and multi-head self-attention mechanisms to retrieve information from historical time-steps. TransAnomaly \cite{yuan2021transanomaly} merges a modified ViViT \cite{arnab2021vivit} with U-Net to augment predictions with richer temporal and global knowledge. However, these VAD transformers \cite{feng2021convolutional,yuan2021transanomaly} expose several limitations when compared to transformers applied in other vision tasks. First, they significantly alter the architecture and behavior of the traditional transformer,  making it hard to quantify the actual contributions of internal transformer components.  Second, these VAD transformers require joint training with discriminators in Generative Adversarial Networks (GANs) \cite{goodfellow2020generative}, leading to additional complexity in hyperparameter tuning and computations. Third and most importantly, despite these endeavors, these VAD transformers continue to underperform many non-transformer methods \cite{doshi2020continual,doshi2022modular,wu2022self,leroux2022multi,deng2023bi,samuel2021svd,wang2020cluster} when evaluated on standard benchmarks. Therefore, we feel obliged to investigate how vision transformers can genuinely thrive and contribute to the VAD domain. To this end, we propose a convolutional temporal transformer that can adeptly handle high-dimensional frames and capture long-range inter-frame dependencies without relying on GANs.  Meanwhile, we cautiously retain much of the original transformer's architecture and behavior, aiming to unleash its full potential in the VAD domain. Compared to prior VAD transformers, ours attains highly competitive performance on VAD benchmarks while 
adopting a simpler one-stage training process.

 \subsection{ConvLSTM}
The ConvLSTM \cite{shi2015convolutional}, a convolutional variant of the fully-connected LSTM (FC-LSTM) \cite{hochreiter1997long}, is another highly representative and widely adopted technique for prediction tasks in computer vision, having prevailed before vision transformers took over. Unlike the limited use of vision transformers, ConvLSTMs have already been thoroughly studied as a core component in many prediction-based VAD works \cite{luo2017remembering,wang2018abnormal,lee2019bman, song2019learning, shen2022video}, proving their suitability for the VAD domain.  Compared to vision transformers \cite{feng2021convolutional,yuan2021transanomaly}, ConvLSTMs do not require feature embedding or dimensionality reduction, which can potentially compromise local image details.  Instead, their fixed receptive fields, defined by convolutional kernels, focus on local spatial dependencies \cite{li2018videolstm}, making ConvLSTMs adept at capturing low-level, fine-grained features like edges and textures. However, this local focus restricts their ability to model higher-level, hierarchical features, unless multiple ConvLSTMs are stacked, which drastically increases computational complexity due to their recurrent nature\cite{shi2015convolutional, wang2018abnormal}. In addition, their inherent first-order Markovian nature impedes information exchange across non-adjacent time-steps \cite{su2020convolutional}, reducing the utilization of long-range temporal dependencies. Recognizing these strengths and limitations, we are the first to propose that the ConvLSTMs' local focus and the vision transformers' long-range modeling can seamlessly complement each other for improved prediction performance. Thus, we devise a layer-interactive ConvLSTM bridge consisting of two ConvLSTMs. This bridge is designed to exclusively transmit short-range local features across layers and time-steps, thus assisting our transformer decoder in capturing fine-grained details. Its compact 2-layer structure alleviates the computational burden associated with recurrence, while retaining the capability to capture shallow spatio-temporal dynamics. Moreover, its distinctive inter-layer interaction strengthens the exploitation of hidden states between the two ConvLSTMs.

\begin{figure*}[t]

\vspace{-4mm}
\centering
\includegraphics[width=\textwidth]{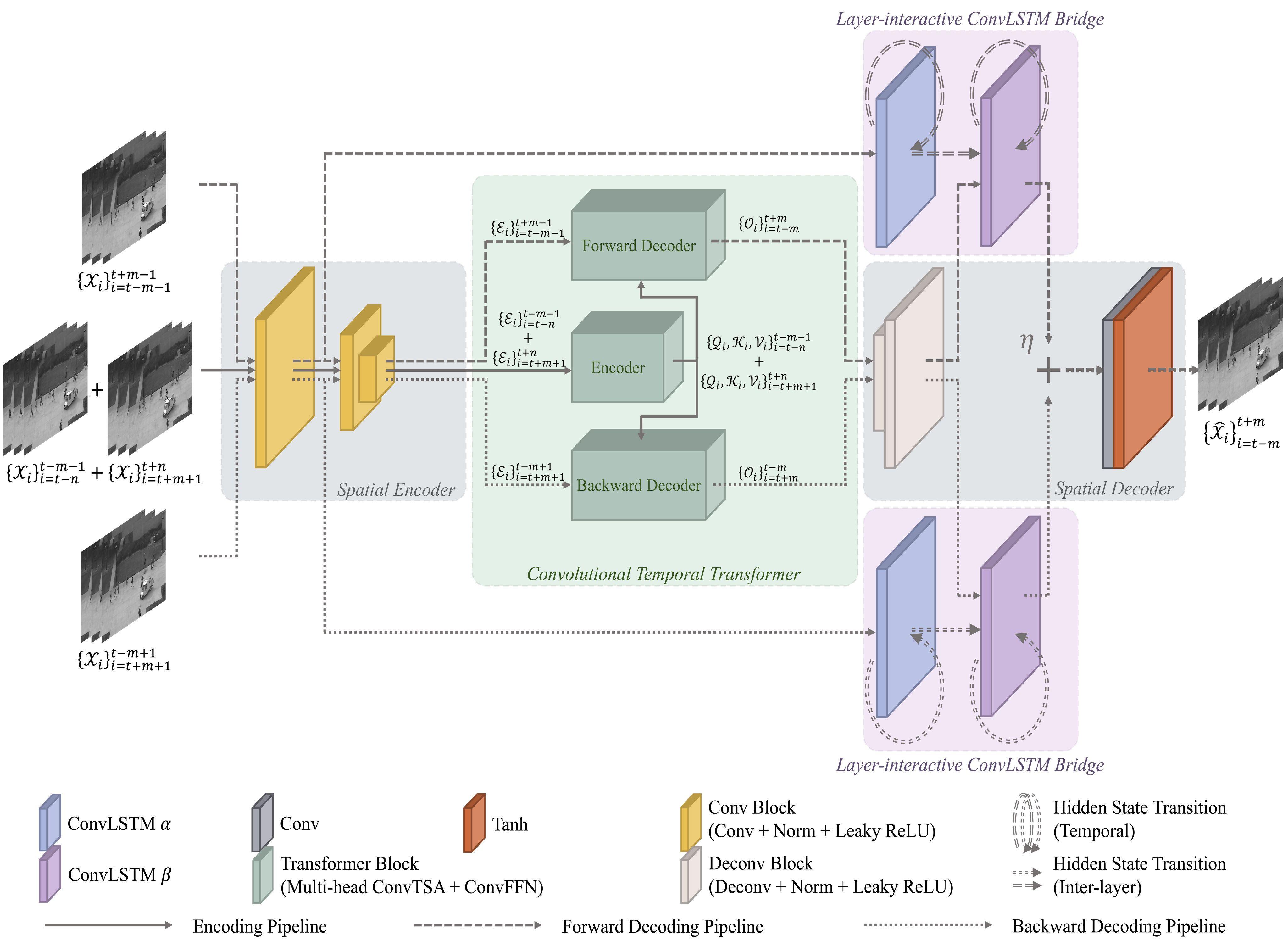}
\caption{Overview of the proposed framework. It includes an encoding pipeline and two decoding pipeline (forward and backward). The encoding pipeline consists of a spatial encoder and a Convolutional Temporal Transformer (ConvTTrans) encoder. Either the forward or backward decoding pipeline comprises the same spatial encoder from the encoding pipeline, a ConvTTrans decoder, a layer-interactive ConvLSTM bridge, and a spatial decoder.} 
\label{fig:modeloverview}
      \vspace{-4mm}
\end{figure*}

\section{Proposed Framework}
\label{sec:profra}

Figure~\ref{fig:modeloverview} illustrates the overall architecture of our bi-directional framework. It mainly consists of an encoding pipeline (solid lines) and two symmetrical decoding pipelines (forward and backward, shown with different dashed lines). The encoding pipeline extracts contextual knowledge from context frames simultaneously, leveraging the parallelization of the transformer encoder. In contrast, the decoding pipelines process target frames, query contextual knowledge, and generate predictions one frame at a time due to the sequential nature of the transformer decoder and ConvLSTMs \cite{vaswani2017attention,shi2015convolutional}. For instance, in the forward decoding pipeline, the initial target frame predicts the second frame, the second frame predicts the third, and so on. Notably, the forward and backward pipelines generate predictions for the same frames but from opposite temporal directions. Finally, predictions from both decoding pipelines are merged to form the final predictions for anomaly inference. This bi-directional framework efficiently alleviates the instability of anomaly scores and the high rate of false alarms commonly seen in uni-directional frameworks \cite{deng2023bi,lee2019bman,shen2022video}. Combining the components depicted in Fig.~\ref{fig:modeloverview}, the process can be detailed as follows. 

Let $\mathcal{X}_{i}$ denote the {\it (i)}-th frame in the video, and $\{\mathcal{X}_i\}_{i=t-n}^{t}$ denote the video clip containing from the {\it (t-n)}-th frame to the {\it (t)}-th frame. Given a video clip $\{\mathcal{X}_i\}_{i=t-m-1}^{t+m+1}$, it is divided into four parts: context clips $\{\mathcal{X}_i\}_{i=t-n}^{t-m-1}$ and $\{\mathcal{X}_i\}_{i=t+m+1}^{t+n}$, a forward target clip $\{\mathcal{X}_i\}_{i=t-m-1}^{t+m-1}$, and a backward target clip $\{\mathcal{X}_i\}_{i=t+m+1}^{t-m+1}$, where $n > m + 1>1$. 

In the encoding pipeline, a spatial encoder embeds the context frames $\{\mathcal{X}_i\}_{i=t-n}^{t-m-1}$ and $\{\mathcal{X}_i\}_{i=t+m+1}^{t+n}$ into their respective spatial feature maps $\{\mathcal{E}_i\}_{i=t-n}^{t-m-1}$ and $\{\mathcal{E}_i\}_{i=t+m+1}^{t+n}$. From these feature maps, a  convolutional transformer encoder then simultaneously extracts contextual knowledge $\{\mathcal{Q}_i, \mathcal{K}_i, \mathcal{V}_i\}_{i=t-n}^{t-m-1}$ and $\{\mathcal{Q}_i, \mathcal{K}_i, \mathcal{V}_i\}_{i=t+m+1}^{t+n}$. 

In the forward decoding pipeline, the same spatial encoder embeds the target frames $\{\mathcal{X}_i\}_{i=t-m-1}^{t+m-1}$ into feature maps $\{\mathcal{E}_i\}_{i=t-m-1}^{t+m-1}$. Subsequently, a convolutional transformer decoder leverages these feature maps and queries the extracted contextual knowledge, $\{\mathcal{Q}_i, \mathcal{K}_i, \mathcal{V}_i\}_{i=t-n}^{t-m-1}$ and $\{\mathcal{Q}_i, \mathcal{K}_i, \mathcal{V}_i\}_{i=t+m+1}^{t+n}$, to generate intermediate predictions $\{\mathcal{O}_i\}_{i=t-m}^{t+m}$ sequentially from time-step {\it (t-m)} to {\it (t+m)}. Specifically,  $\mathcal{E}_{t-m-1}$ is used to predict $\mathcal{O}_{t-m}$, $\mathcal{E}_{t-m}$ is used to predict $\mathcal{O}_{t-m+1}$, $\mathcal{E}_{t-m+1}$ is used to predict $\mathcal{O}_{t-m+2}$, and so on, until $\mathcal{E}_{t+m-1}$ predicts $\mathcal{O}_{t+m}$. 

Similarly, in the backward decoding pipeline, the spatial encoder embeds another target clip $\{\mathcal{X}_i\}_{i=t+m+1}^{t-m+1}$ into $\{\mathcal{E}_i\}_{i=t+m+1}^{t-m+1}$, and a second transformer decoder queries the contextual knowledge and predicts in the reverse temporal direction, producing $\{\mathcal{O}_i\}_{i=t+m}^{t-m}$ sequentially from time-step {\it (t+m)} to {\it (t-m)}. Specifically,  $\mathcal{E}_{t+m+1}$ is used to predict $\mathcal{O}_{t+m}$, $\mathcal{E}_{t+m}$ is used to predict $\mathcal{O}_{t+m-1}$, $\mathcal{E}_{t+m-1}$ is used to predict $\mathcal{O}_{t+m-2}$, and so on, until $\mathcal{E}_{t-m+1}$ predicts $\mathcal{O}_{t-m}$. 

Both forward and backward predictions, $\{\mathcal{O}_i\}_{i=t-m}^{t+m}$ and $\{\mathcal{O}_i\}_{i=t+m}^{t-m}$, are refined by their respective layer-interactive ConvLSTM bridges during the upsampling process by the spatial decoder. As shown in Fig.\ref{fig:modeloverview}, these ConvLSTM bridges connect the spatial encoder and decoder, efficiently propagating low-level features across multiple time-steps and layers. After the ConvLSTM bridges, the refined predictions for the same frame from both decoding pipelines are linearly summed with a weighting factor $\eta$ to construct the final predictions $\{\hat{\mathcal{X}}_i\}_{i=t-m}^{t+m}$ for anomaly detection. During inference, prediction discrepancies are calculated through a combined objective function to estimate abnormalities. 

There are three critical aspects to emphasize regarding the framework's structure:

\begin{enumerate}
\item As mentioned earlier, the forward and backward decoding pipelines have their own transformer decoders and layer-interactive ConvLSTM bridges. However, both pipelines, along with the encoding pipeline, share the same spatial encoder and decoder to ensure the framework's stability and efficiency.

\item The contextual knowledge, $\{\mathcal{Q}_i, \mathcal{K}_i, \mathcal{V}_i\}_{i=t-n}^{t-m-1}$ and $\{\mathcal{Q}_i, \mathcal{K}_i, \mathcal{V}_i\}_{i=t+m+1}^{t+n}$, extracted from two context clips, $\{\mathcal{X}_i\}_{i=t-n}^{t-m-1}$ and $\{\mathcal{X}_i\}_{i=t+m+1}^{t+n}$, is fed into each decoding pipeline. This operation assumes that our convolutional transformer can associate non-consecutive frames, akin to how NLP transformers relate non-consecutive word tokens from sentences. 

\item The decoding pipeline, which combines the convolutional transformer decoder with the layer-interactive ConvLSTM bridge, leverages target frames to predict subsequent frames sequentially. That is, the decoding pipeline always uses the ground truth from the last time-step as input for the current time-step's prediction. This sequential prediction technique, known as teacher forcing \cite{goodfellow2016deep}, is well-established in NLP transformers and LSTMs for language translation tasks.  


\end{enumerate}

In the following sections, we elaborate on these modules: spatial encoder/decoder (Section~\ref{sec:featureembedding}); convolutional temporal transformer (Section~\ref{sec:ConvTTrans}); and layer-interactive ConvLSTM bridge (Section~\ref{subsec:bridge}). For clarity, all modules are explained in the context of the forward (uni-directional) prediction scenario, while the backward scenario can be easily inferred by reversing the order of frames. In addition, we introduce a scheme to prevent information leakage when integrating the transformer decoder and ConvLSTM bridge for sequential prediction (Section~\ref{subsec:infoleak}), followed by a description of the anomaly inference process (Section~\ref{subsec:scoring}).

\subsection{Spatial Encoder/Decoder}
\label{sec:featureembedding}



As depicted in Fig.~\ref{fig:modeloverview}, our spatial encoder, which consists of three classic convolutional blocks, maps each frame $\mathcal{X}_i$ independently to its corresponding spatial feature map $\mathcal{E}_i$. On the other hand, our spatial decoder roughly mirrors the encoder's structure in reverse order, with convolutional layers replaced by deconvolutional layers. It takes the output of the transformer decoder, $\mathcal{O}_{i}$, to construct the predicted frame $\widehat{\mathcal{X}}_{i}$, while incorporating refinements from the layer-interactive ConvLSTM bridge. Both the spatial encoder and decoder only exploit spatial correlations within individual frames, leaving the convolutional transformer and layer-interactive ConvLSTM bridge responsible for managing temporal dynamics.

\subsection{Convolutional Temporal Transformer (ConvTTrans)}
\label{sec:ConvTTrans}

 \begin{figure}

  \centering
  \includegraphics[width=0.75\linewidth]{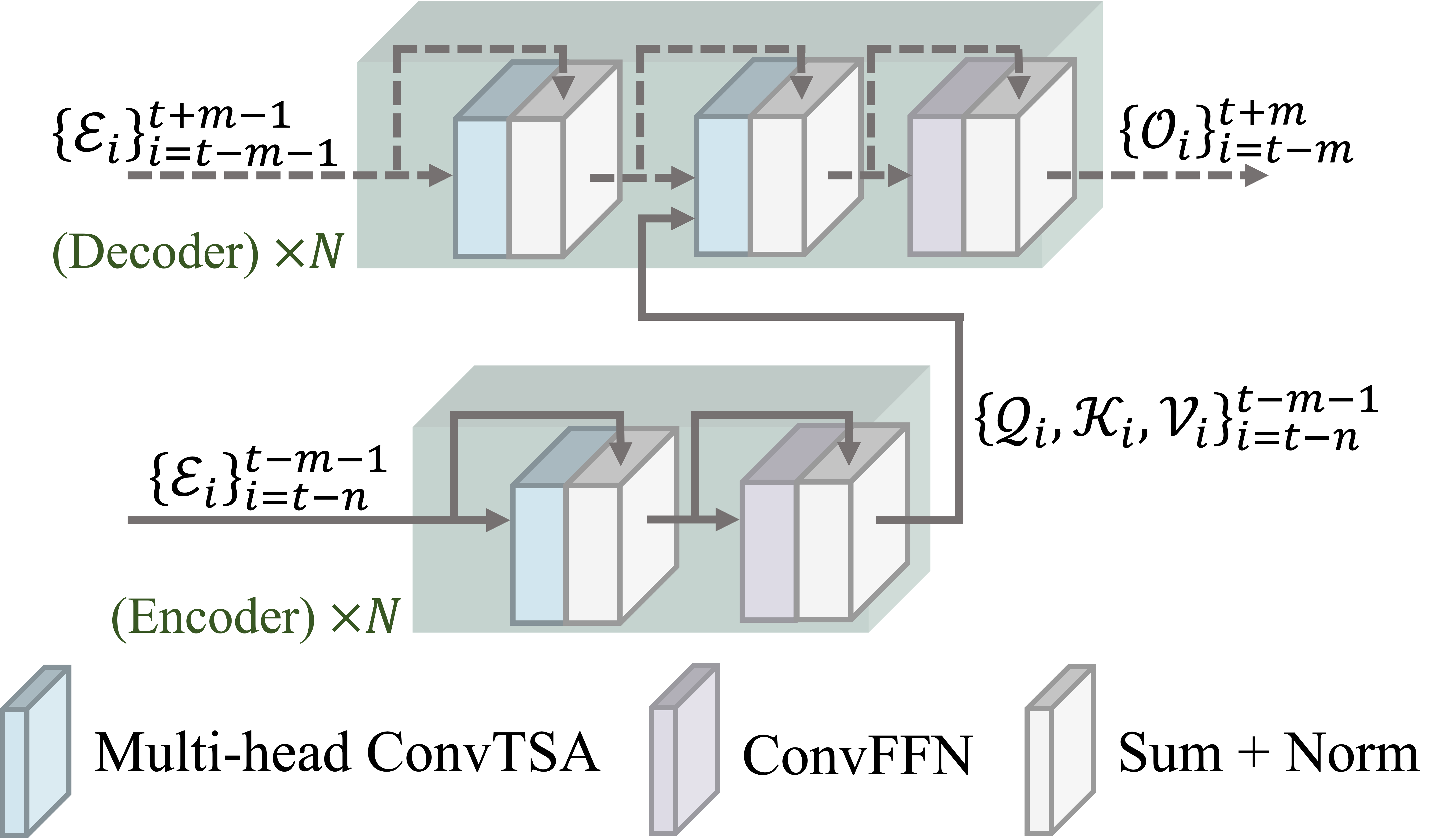}
   \caption{The encoder-decoder architecture of the Convolutional Temporal Transformer (ConvTTrans). Only the encoder and forward decoder are visualized; the backward decoder behaves similarly to the forward decoder but with different inputs.}
   
   \label{fig:forward_convttrans}
\end{figure}

To capture long-range dependencies among frames, we propose the ConvTTrans based on the vanilla transformer \cite{vaswani2017attention}. 
As shown in Fig.~\ref{fig:forward_convttrans}, the ConvTTrans encoder processes all context feature maps $\{\mathcal{E}_i\}_{i=t-n}^{t-m-1}$ in parallel to extract contextual knowledge $\{\mathcal{Q}_i, \mathcal{K}_i, \mathcal{V}_i\}_{i=t-n}^{t-m-1}$. The ConvTTrans decoder then queries this knowledge and takes the target feature maps $\{\mathcal{E}_i\}_{i=t-m-1}^{t+m-1}$ to yield the predictions $\{\mathcal{O}_i\}_{i=t-m}^{t+m}$ sequentially. Both the ConvTTrans encoder and decoder are designed as a cascade of $N$ identical blocks to extract high-level representations. Specifically, each encoder block consists of a multi-head Convolutional Temporal Self-attention (ConvTSA) and a Convolutional Feed-forward Network (ConvFFN); each decoder block comprises two multi-head ConvTSAs and a ConvFFN. In addition, a residual connection is introduced after each multi-head ConvTSA or ConvFFN, followed by channel-wise spatial normalization.   The following sections detail the architecture of each module.

 \begin{figure}
  \centering
  \includegraphics[width=\linewidth]{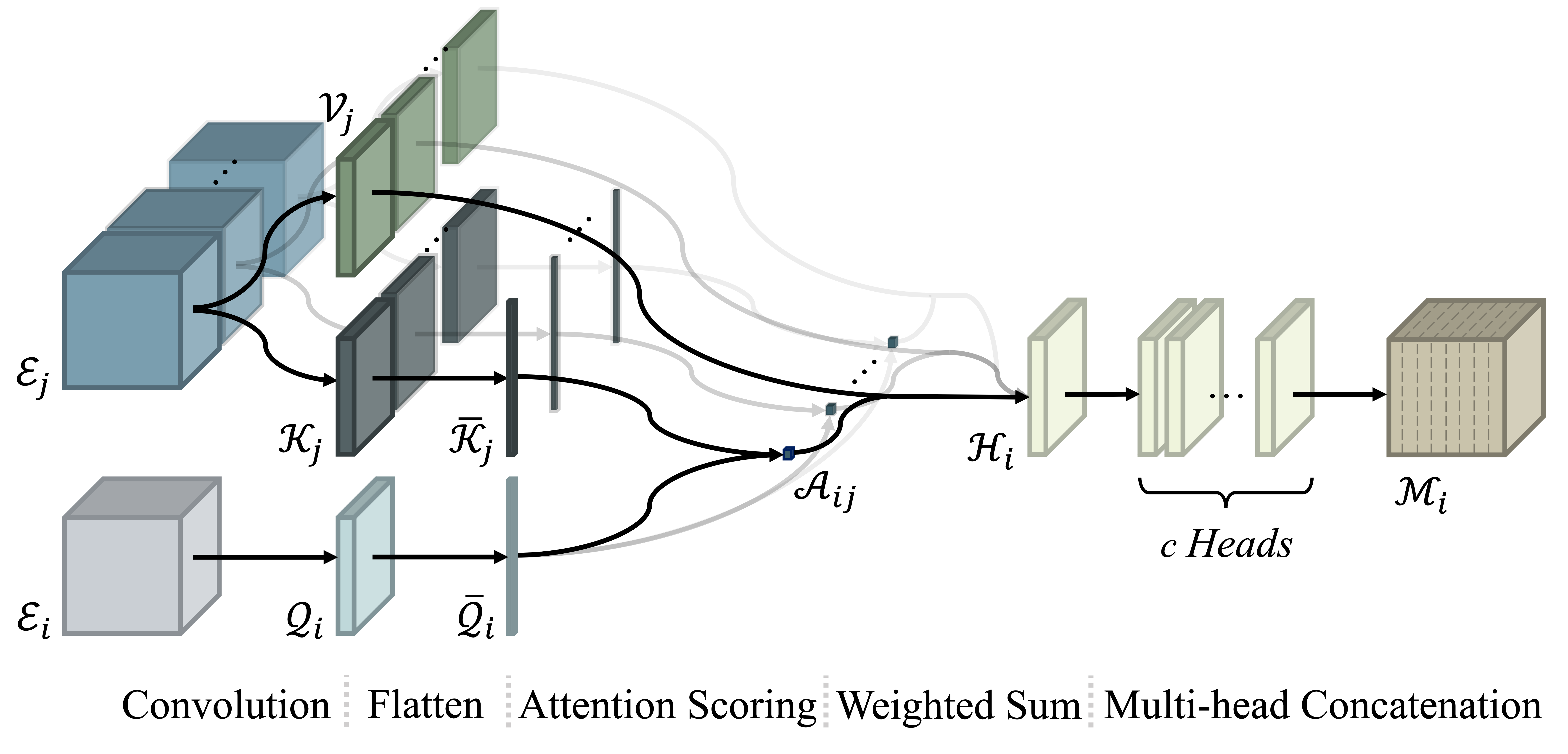}
   \caption{Illustration of the multi-head ConvTSA. The attention-based representation $\mathcal{H}_{i}$ between frame $\mathcal{X}_{i}$ (embedded into $\mathcal{E}_{i}$) and every other frame $\mathcal{X}_{j}$ in the clip (embedded into  $\mathcal{E}_{j}$) is simultaneously calculated in multiple heads. Note that the actual number of $\{\mathcal{E}_j\}$ depends on the clip length.} 

   \label{fig:multihead}
\end{figure}

{\bf Convolutional Temporal Self-attention (ConvTSA).} This module aims to associate all time-steps within a clip to create a temporal attention-based representation for each frame. As shown in Fig.~\ref{fig:multihead}, given the spatial feature map $\mathcal{E}_i$ of frame $\mathcal{X}_i$, we employ convolution operations to generate its query $\mathcal{Q}_i$, key $\mathcal{K}_i$, and value  $\mathcal{V}_i$. 
Unlike ViT-based approaches \cite{dosovitskiy2020image,arnab2021vivit}, which directly adopt the multiplicative operations from NLP transformers, our use of convolutions can retain significant spatial information and involve fewer learnable parameters.

After completing the above operations for each frame, the temporal attention score $\mathcal{A}_{ij}$ is calculated to determine how relevant a particular frame $\mathcal{X}_{i}$ is with respect to {\it every} other frame $\mathcal{X}_{j}$ in the clip. To this end, we flatten frame $\mathcal{X}_{i}$'s   $\{\mathcal{Q}_{i}, \mathcal{K}_{i}\}$ to $\{\Bar{\mathcal{Q}}_{i}, \Bar{\mathcal{K}}_{i} \}$ and frame $\mathcal{X}_{j}$'s   $\{\mathcal{Q}_{j}, \mathcal{K}_{j}\}$ to $\{\Bar{\mathcal{Q}}_{j}, \Bar{\mathcal{K}}_{j} \}$, 
and then score every frame $\mathcal{X}_{j}$ against frame $\mathcal{X}_{i}$ as follows:
\begin{equation}
\mathcal{A}_{ij} = \frac{\Bar{\mathcal{Q}}_i \cdot \Bar{\mathcal{K}}_j}{\sqrt{d}}, 
\label{eq:tsfmatt}
\end{equation}
where 
the dot-product is conducted between $\Bar{\mathcal{Q}}_i$ and $\Bar{\mathcal{K}}_j$ as in \cite{vaswani2017attention}; and the scaling factor $d$ is set to the length of $\Bar{\mathcal{Q}}_i$ (or $\Bar{\mathcal{K}}_j$). Consequently, score $\mathcal{A}_{ij}$ obtained from Eq.~\ref{eq:tsfmatt} is a scalar value. Since each frame $\mathcal{X}_{j}$ generates an $\mathcal{A}_{ij}$ with respect to frame $\mathcal{X}_{i}$, we subsequently normalize $\mathcal{A}_{ij}$ to $\mathcal{A}_{ij}^{\prime}$ by performing $\operatorname { Softmax }\left(\cdot\right)$ on all the scores $\{\mathcal{A}_{ij}\}$ within the clip, as follows:
\begin{equation}
\mathcal{A}_{ij}^{\prime} = \operatorname { Softmax } (\mathcal{A}_{ij})_j = \frac{e^{\mathcal{A}_{ij}}}{\sum_{k} e^{\mathcal{A}_{ik}}}.
\label{eq:softmax}
\end{equation}

Finally, ConvTSA outputs an attention-based representation $\mathcal{H}_{i}$ for frame $\mathcal{X}_{i}$. This representation $\mathcal{H}_{i}$ encapsulates the information from all frames $\{\mathcal{X}_{j}\}$ in the clip to which frame $\mathcal{X}_{i}$ attends. To this end, we utilize the normalized attention score $\mathcal{A}_{ij}^{\prime}$ to weigh the value $\mathcal{V}_j$ of  frame $\mathcal{X}_{j}$ and 
then sum them up as follows:
\begin{equation}
\mathcal{H}_{i}=\sum_{j} \mathcal{A}_{ij}^{\prime} \mathcal{V}_j.
\label{eq:outatt}
\end{equation}

{\bf Multi-head.} This mechanism provides the above ConvTSA with richer representations across multiple sub-spaces. As depicted in Fig.~\ref{fig:multihead}, ConvTSA is executed  in $c$ different {\it heads} simultaneously, each having its unique learnable parameters. The outputs for frame $\mathcal{X}_{i}$ from all heads, denoted as $\mathcal{H}_{i}^{(1)},\mathcal{H}_{i}^{(2)}, \ldots, \mathcal{H}_{i}^{(c)}$, are then concatenated along the channel dimension to form the multi-head attention $\mathcal{M}_{i}$:
\begin{equation}
\mathcal{M}_{i}=\operatorname { Concat }(\underbrace{\mathcal{H}_{i}^{(1)},\mathcal{H}_{i}^{(2)}, \ldots, \mathcal{H}_{i}^{(c)}}_\textit{c heads}).
\label{eq:concatatt}
\end{equation}


Note that the second multi-head ConvTSA in each decoder block conducts the context query. Instead of computing multi-head attentions within clips, it asymmetrically scores the target frames $\{\mathcal{X}_i\}_{i=t-m-1}^{t+m-1}$   against the context frames $\{\mathcal{X}_i\}_{i=t-n}^{t-m-1}$ by utilizing contextual knowledge $\{\mathcal{Q}_i, \mathcal{K}_i, \mathcal{V}_i\}_{i=t-n}^{t-m-1}$, which is generated by the final ConvTTrans encoder block in Fig.~\ref{fig:forward_convttrans}. 

{\bf Convolutional Feed-forward Network (ConvFFN).} This module, depicted in Fig.~\ref{fig:forward_convttrans} and derived from the position-wise feed-forward network (PFFN) \cite{vaswani2017attention}, is designed to render the multi-head attention $\mathcal{M}_{i}$ more compatible with the subsequent layer. To this end, we replace the fully-connected layers of PFFN \cite{vaswani2017attention} with convolutions to match $\mathcal{M}_{i}$'s dimensionality, while maintaining the use of Leaky ReLU activations to produce non-linearities.


\subsection{Layer-interactive ConvLSTM (LI-ConvLSTM) Bridge}
\label{subsec:bridge}

\begin{figure}[t]

  \centering
  \includegraphics[width=\linewidth]{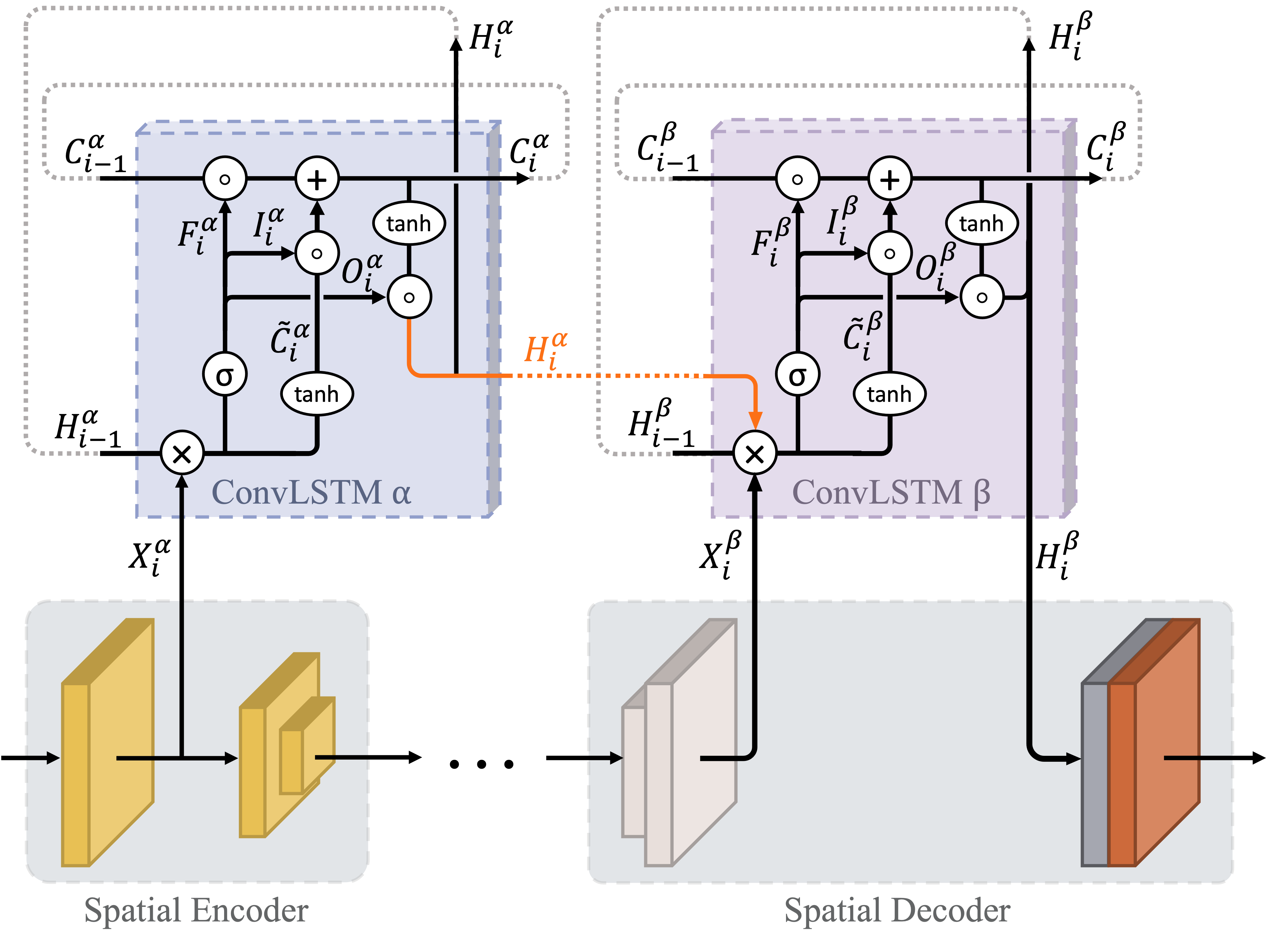}
  
   
   \caption{Architecture of the layer-interactive ConvLSTM bridge and its connection with the spatial encoder and decoder. 
   ConvLSTM $\alpha$  and ConvLSTM $\beta$ are the traditional and modified ConvLSTMs, respectively; $H^{\alpha}_{i}$ and orange dashed lines indicate the inter-layer transition of hidden state from ConvLSTM $\alpha$ to ConvLSTM $\beta$; and grey dashed lines symbolize the standard temporal transition of hidden and cell states within ConvLSTMs \cite{shi2015convolutional}. Each module corresponds to the module depicted in the same color in Fig.~\ref{fig:modeloverview}.}
   \label{fig:bridge}
\end{figure}

The spatial encoder discussed in Section~\ref{sec:featureembedding}  may result in an irreversible loss of low-level features, thus hampering the accurate prediction of fine details. Inspired by PredRNN \cite{wang2017predrnn}, we introduce a unique LI-ConvLSTM bridge to efficiently transfer these low-level features across time and layers. 
As depicted in Fig.~\ref{fig:bridge}, the bridge is constructed as a cascade of two similar ConvLSTM layers at the same scale, referred to as ConvLSTM $\alpha$ and ConvLSTM $\beta$, with their respective components denoted by the superscripts $^\alpha$ or $^\beta$.

Our LI-ConvLSTM bridge is positioned immediately after the first layer of the spatial encoder and extends to just before the penultimate layer of the spatial decoder. This design is based on the observation that low-level features are typically located in shallow layers. Specifically, ConvLSTM $\alpha$ takes the $X^{\alpha}_i$ from the first layer of the spatial encoder at time-step $i$ as its input and generates its hidden state $H^{\alpha}_i$. ConvLSTM $\beta$ then uses both the hidden state $H^{\alpha}_i$ from ConvLSTM $\alpha$ and the $X^{\beta}_i$ from the mid-layer of the spatial decoder to generate its hidden state $H^{\beta}_i$. The hidden state $H^{\beta}_i$ is subsequently utilized by the following layers of the spatial decoder for frame prediction. Meanwhile, all hidden and cell states in the bridge, namely $H^{\alpha}_i$, $H^{\beta}_i$, $C^{\alpha}_i$, $C^{\beta}_i$,  transit temporally within their respective modules, ConvLSTM $\alpha$ or $\beta$, in a manner consistent with standard ConvLSTMs \cite{shi2015convolutional} (see grey dashed lines in Fig.~\ref{fig:bridge}).

ConvLSTM $\alpha$, which connects to the spatial encoder, resembles the original ConvLSTM described in \cite{shi2015convolutional}. Specifically, it shares the same architecture as the LSTM module commonly used in the NLP domain, but replaces all matrix multiplications with convolution operations. Thus, ConvLSTM $\alpha$ can be formulated as follows:
\begin{align}
F^{\alpha}_i & = \sigma(W_F \otimes [H^{\alpha}_{i-1},  X^{\alpha}_i] + b_F), \label{eq:forget_gate} \\
I^{\alpha}_i & = \sigma(W_I \otimes [H^{\alpha}_{i-1}, X^{\alpha}_i] + b_I), \label{eq:input_gate} \\
\tilde{C}^{\alpha}_i & = \tanh(W_{\tilde{C}} \otimes [H^{\alpha}_{i-1},  X^{\alpha}_i] + b_{\tilde{C}}), \label{eq:candidate_cell} \\
C^{\alpha}_i & = F^{\alpha}_i \circ C^{\alpha}_{i-1} + I^{\alpha}_i \circ \tilde{C}^{\alpha}_i, \label{eq:cell_state} \\
O^{\alpha}_i & = \sigma(W_O \otimes [H^{\alpha}_{i-1}, X^{\alpha}_i] + b_O), \label{eq:output_gate} \\
H^{\alpha}_i & = O^{\alpha}_i \circ \tanh(C^{\alpha}_i), \label{eq:hidden_state}
\end{align}
where $X^{\alpha}_i$ is the input to ConvLSTM $\alpha$ at time-step $i$, derived from the preceding layer of the spatial encoder; $H^{\alpha}_{i}$ is the hidden state at time-step $i$; $H^{\alpha}_{i-1}$ is the hidden state that transits within ConvLSTM $\alpha$ from time-step $(i-1)$; 
$\tilde{C}^{\alpha}_i$, $C^{\alpha}_i$, $I^{\alpha}_i$, $F^{\alpha}_i$, and $O^{\alpha}_i$ denote the candidate cell state, cell state, input gate, forget gate, and output gate, respectively; and $\sigma(\cdot)$, $\circ$, and $\otimes$ denote the Sigmoid, Hadamard, and convolution operations, respectively.
 
It is clear that ConvLSTM $\alpha$ conducts the standard temporal transition of hidden states \cite{shi2015convolutional}, where these states are trapped within the layer and can only be transferred across time-steps. Researchers today argue that this mechanism may inhibit the efficient exchange and utilization of spatio-temporal knowledge across non-adjacent layers \cite{wang2017predrnn,wang2018predrnn++,8953605}. To address this issue, we introduce a modified ConvLSTM $\beta$, which follows ConvLSTM $\alpha$ and connects to the spatial decoder, to facilitate an additional inter-layer transition of hidden states. Accordingly, Equations~\eqref{eq:forget_gate}, \eqref{eq:input_gate}, \eqref{eq:candidate_cell}, and \eqref{eq:output_gate} are modified for ConvLSTM $\beta$, and ConvLSTM $\beta$ can be calculated as follows: 
\begin{align}
F^{\beta}_i & = \sigma(W_F \otimes [H^{\beta}_{i-1}, \boldsymbol{H^{\alpha}_i}, X^{\beta}_i] + b_F), \label{eq:forget_gate_beta} \\
I^{\beta}_i & = \sigma(W_I \otimes [H^{\beta}_{i-1}, \boldsymbol{H^{\alpha}_i}, X^{\beta}_i] + b_I), \label{eq:input_gate_beta} \\
\tilde{C}^{\beta}_i & = \tanh(W_{\tilde{C}} \otimes [H^{\beta}_{i-1}, \boldsymbol{H^{\alpha}_i},  X^{\beta}_i] + b_{\tilde{C}}), \label{eq:candidate_cell_beta} \\
C^{\beta}_i & = F^{\beta}_i \circ C^{\beta}_{i-1} + I^{\beta}_i \circ \tilde{C}^{\beta}_i, \label{eq:cell_state_beta} \\
O^{\beta}_i & = \sigma(W_O \otimes [H^{\beta}_{i-1}, \boldsymbol{H^{\alpha}_i}, X^{\beta}_i] + b_O), \label{eq:output_gate_beta} \\
H^{\beta}_i & = O^{\beta}_i \circ \tanh(C^{\beta}_i). \label{eq:hidden_state_beta}
\end{align}

Here, unlike ConvLSTM $\alpha$, the input $X^{\beta}_i$ to ConvLSTM $\beta$ is from the preceding layer of the spatial decoder; and $H^{\beta}_{i-1}$ is the hidden state that transits within ConvLSTM $\beta$ from time-step $(i-1)$. While ConvLSTM $\beta$ largely retains the original structure of ConvLSTM, it exclusively incorporates the hidden state $H^{\alpha}_i$ from ConvLSTM $\alpha$ to compute the candidate cell state $\tilde{C}^{\beta}_i$, input gate $I^{\beta}_i$, forget gate $F^{\beta}_i$, and output gate $O^{\beta}_i$.


Unlike densely-stacked ConvLSTM frameworks \cite{wang2018abnormal, lee2019bman,shen2022video}, our LI-ConvLSTM bridge is intentionally tuned to a compact two-layer structure, considering the simplicity of modelling low-level features and the inefficiency of performing excessive recurrence. Our ablation study further reveals that this bridge not only enables direct inter-layer interaction of low-level features, akin to a residual connection, but also preserves the ConvLSTM's capacity to capture shallow spatio-temporal dynamics. This leads to its superiority over traditional residual connection approaches. 

\subsection{Scheme against Information Leakage}
\label{subsec:infoleak}
After encoding all target frames through the spatial encoder (Section~\ref{sec:featureembedding}), our ConvTTrans decoder and LI-ConvLSTM bridge adopt different mechanisms to prevent information leakage during sequential prediction.  In the ConvTTrans decoder, a triangular mask \cite{vaswani2017attention} is applied to the first multi-head ConvTSA of each block. This mask ensures that a particular frame $\mathcal{X}_{i}$ only attends to preceding (subsequent) frames $\{\mathcal{X}_{j}\}$ during the forward (backward) prediction.  On the other hand,  the LI-ConvLSTM bridge, being a typical recurrent structure, inherently has the capability to process one frame at a time without ``peeking" ahead.

\subsection{Anomaly Inference}
\label{subsec:scoring}

During the training phase, we leverage a combined objective function to evaluate the predictions at both pixel and perceptual levels. The pixel-level abnormality is quantified by the standard Mean Absolute Error (MAE), denoted as $\mathcal{L}_{mae}(\mathcal{X}_i,\widehat{\mathcal{X}}_i)$, which measures the individual pixel-wise discrepancies between frame $\mathcal{X}_i$ and its prediction $\widehat{\mathcal{X}}_i$.
%
The perceptual-level abnormality is estimated by the Structural Similarity Index Measure (SSIM) \cite{wang2004image}, denoted as  $\mathcal{L}_{ssim}(\mathcal{X}_i,\widehat{\mathcal{X}}_i)$, which focuses on perceptually-motivated properties \cite{Sgaard2016ApplicabilityOE,wang2011information} such as frame structure, contrast, and luminance.
%
The final objective function $\mathcal{L}(\mathcal{X}_i,\widehat{\mathcal{X}}_i)$ is formulated as a weighted sum of the MAE and SSIM losses:
\begin{equation}
\mathcal{L}(\mathcal{X}_i,\widehat{\mathcal{X}}_i)= \mathcal{L}_{ssim}(\mathcal{X}_i,\widehat{\mathcal{X}}_i)+\lambda ( W_{g} \otimes \mathcal{L}_{mae}(\mathcal{X}_i,\widehat{\mathcal{X}}_i)),
\label{eq:mixloss}
\end{equation}
where the weighting factor $\lambda$ determines the relative importance of these two metrics. In practice, SSIM is locally calculated by sliding a Gaussian weighing filter over the entire image \cite{wang2003multiscale,pytorch-ignite}. To maintain consistency between these two metrics, we employ the same Gaussian filter, which is parameterized by $W_{g}$ in Eq.~\ref{eq:mixloss}, to weigh pixels and perform local MAE computations, resembling the SSIM process.


During the inference phase, we continuously extract a sequence $\{\mathcal{X}_i\}_{i=t-n}^{t+n}$ from the testing video stream, dividing it into the forward target clip $\{\mathcal{X}_i\}_{i=t-m-1}^{t+m-1}$, the backward target clip $\{\mathcal{X}_i\}_{i=t+m+1}^{t-m+1}$, and context clips $\{\mathcal{X}_i\}_{i=t-n}^{t-m-1}$ and $\{\mathcal{X}_i\}_{i=t+m+1}^{t+n}$. Our trained framework then leverages these clips to generate the prediction sequence $\{\hat{\mathcal{X}}_i\}_{i=t-m}^{t+m}$. With the bi-directional framework's symmetry in mind, we specifically focus on the {\it midmost} frame $\mathcal{X}_{t}$ and its prediction $\widehat{\mathcal{X}}_{t}$, and utilize  Eq.~\ref{eq:mixloss} to produce an anomaly score $\mathcal{L}(\mathcal{X}_{t},\widehat{\mathcal{X}}_{t})$. This score is subsequently normalized using  min-max normalization \cite{cai2021appearance,doshi2023towards,gong2019memorizing,nguyen2019anomaly,Park2022FastAnoFA,park2020learning} for effective comparisons with other works. A higher value of $\mathcal{L}(\mathcal{X}_{t},\widehat{\mathcal{X}}_{t})$ indicates a greater degree of abnormality in $\mathcal{X}_{t}$.


\section{Experiments}
\label{sec:experi}

\subsection{Benchmarks} 
We evaluate our framework on four widely-used public benchmarks for anomaly detection in videos: UCSD Ped2 \cite{mahadevan2010anomaly}, CUHK Avenue \cite{lu2013abnormal}, ShanghaiTech \cite{luo2017revisit}, and Street Scene \cite{ramachandra2020street}. Using these datasets for evaluation allows us to compare our work with the majority of  state-of-the-art methods. 

{\bf UCSD Ped2 (UCSD).} This dataset contains 16 training videos and 12 test videos, all recorded in grayscale format with a resolution of $240\times360$ pixels. These videos capture a walkway where pedestrians move in parallel to the camera plane. Anomalies in this dataset include bicycles, skateboarders, and vehicles.  While this dataset is widely used in VAD, it features a relatively low resolution and a constrained scene setting.

{\bf CUHK Avenue (Avenue).} This dataset contains 16 training videos and 21 test videos, all recorded in RGB format with a resolution of $360\times640$ pixels. These videos depict an avenue on the CUHK campus with varying pedestrian sizes and crowd densities.  Anomalies in this dataset include loitering, running, and throwing objects. However, this dataset is also limited to a single scene setting despite its popularity in VAD.

{\bf ShanghaiTech (SHTech).} This dataset contains 330 training videos and 107 test videos, all recorded in RGB format with a resolution of $480\times856$ pixels. These videos extensively cover 13 scenes of the ShanghaiTech campus with complex lighting conditions and camera angles.  Anomalies in this dataset include all those found in the previous two datasets, along with interpersonal interactions such as brawling, chasing, and snatching. This dataset is renowned as one of the most challenging and comprehensive VAD datasets available.

{\bf Street Scene (StreetScene).} This dataset contains 46 training videos and 35 test videos, all recorded in RGB format with a
resolution of $720\times1280$ pixels. These videos portray a double-lane street with bike lanes and pedestrian sidewalks during the daytime. In addition to the anomalies shared with other datasets, this dataset uniquely includes traffic violations like pedestrian jaywalking, car turning, and bicycles departing from bike lanes. Due to the variety of anomaly types, this dataset is extremely challenging and, as a result, infrequently explored.

\subsection{Implementation Details}

The length of training samples is defined by setting $n$ to 4 and $m$ to 1 in our 
bi-directional framework in Fig.~\ref{fig:modeloverview}. Instead of predicting adjacent frames, we sample the original clips at an interval of every 3 frames to better discriminate the framework's prediction capacity from its reconstruction capacity. Each training sample therefore consists of 9 evenly-spaced individual frames, forming the forward target clip  $\{\mathcal{X}_{t-6}, \mathcal{X}_{t-3}, \mathcal{X}_{t}\}$, backward target clip $\{\mathcal{X}_{t+6}, \mathcal{X}_{t+3}, \mathcal{X}_{t}\}$, and context clips $\{\mathcal{X}_{t-12}, \mathcal{X}_{t-9}, \mathcal{X}_{t-6}\}$ and $\{\mathcal{X}_{t+6},\mathcal{X}_{t+9}, \allowbreak \mathcal{X}_{t+12}\}$.

Each frame $\mathcal{X}_i$ is normalized to the intensity of $[-1,1]$ and re-sized to $256 \times 256 \times 1$ (UCSD) or $256 \times 256 \times 3$ (Avenue, SHTech, StreetScene). Major hyperparameters are set empirically as follows: the size of feature map $\mathcal{E}_i$ and multi-head attention $\mathcal{M}_i$ is $64 \times 64 \times 64$; the size of query $\mathcal{Q}_i$, key $\mathcal{K}_i$,  value  $\mathcal{V}_i$, and single-head attention $\mathcal{H}_{i}$, is $64 \times 64 \times 8$; the size of each state in the LI-ConvLSTM bridge is $256 \times 256 \times 16$; the head number $c$ is $8$; the ConvTTrans block number $N$ is $5$;  the weighting factor $\lambda$ in the objective function is set to $1$; and the weighting factor $\eta$ for two decoding pipelines is set to $0.75$. The receptive field of ConvTSAs and ConvFFNs is set to $3\times3$, while that of ConvLSTM $\alpha$ and $\beta$ is $5\times5$. 

Our framework is built on PyTorch \cite{paszke2017automatic} and trained end-to-end as an OCC with no pre-training. It is optimized by the Adam \cite{kingma2014adam} algorithm with a batch size of 4 clips and an initial learning rate of 0.001 which decays by $50\%$ once improvement stagnates for 3 epochs. To avoid overfitting, we use $10\%$ of the total clips from the default training set as the validation set, and use the optimal validation loss for early stopping. All experiments are conducted on one NVIDIA RTX A5000.


\subsection{Evaluation Criteria} For frame-level evaluation, following most prior works \cite{doshi2020continual, doshi2022modular,leroux2022multi,deng2023bi,  wu2022self, doshi2023towards, Park2022FastAnoFA, wang2020cluster,yu2020cloze},  we adopt the frame-level Area Under Curve (AUC) of the Receiver Operating Characteristic (ROC) curve as our primary metric. For pixel-level evaluation, we consider the region-based detection criterion (RBDC) and track-based detection criterion (TBDC) proposed by Ramachandra et al. \cite{ramachandra2020street}, which measure the detection accuracy of abnormal regions and tracks, respectively. Meanwhile, we provide comprehensive statistics and visualizations to support the pixel-level analysis. However, RBDC and TBDC also exhibit inherent limitations \cite{doshi2023towards,doshi2022modular,Park2022FastAnoFA} and are rarely used in the literature. Hence, we suggest not drawing any definitive conclusions based on RBDC/TBDC, but instead relying on the more robust frame-level AUC.

\subsection{Comparison with the State-of-the-art (SOTA)}

\begin{table}

  \caption{Frame-level AUC (\%) comparison with State-of-the-art OCC-based methods on UCSD  \cite{mahadevan2010anomaly}, Avenue \cite{lu2013abnormal}, and SHTech \cite{luo2017revisit}. The \textbf{best}, \underline{second-best}, and \uwave{third-best} performances are highlighted.}
  \label{tab:auc}
  \vspace*{-3mm}
  \centering
    \setlength{\tabcolsep}{4pt}
    \subfloat[]{%
  \begin{tabular}{clccc}
    \toprule
    & \textbf{Method} & \textbf{UCSD} & \textbf{Avenue} & \textbf{SHTech} \\
    \midrule
  \multirow{15}{*}{\rotatebox[origin=c]{90}{Single-task}}& Gong et al. \cite{gong2019memorizing} & 94.1 & 83.3 & 71.2\\
     & Park et al. \cite{park2020learning} & 97.0 & 88.5 & 72.8 \\
    & Lu et al. \cite{lu2020few} & 96.2 & 85.8 & 77.9 \\
    & Doshi et al. \cite{doshi2020continual} & \uwave{ 97.8 } & 86.4 & 71.6 \\
    & Rodrigues et al. \cite{rodrigues2020multi} & - & 82.9 & 76.0 \\
    & Wang et al. \cite{wang2020cluster} & - & 87.0 & \uwave{ 79.3 } \\      
    & Samuel et al. \cite{samuel2021svd} & 77.0 & \underline{89.8} & 78.4 \\    
    & Feng et al.\cite{feng2021convolutional} &  97.2 & 85.9 & 77.7 \\
    & Yuan et al.\cite{yuan2021transanomaly} & 96.4 & 87.0 & - \\
    & Park et al. \cite{Park2022FastAnoFA} & 96.3 & 85.3 & 72.2 \\
    & Doshi et al. \cite{doshi2022modular} & 97.2 & 88.7 & 73.6 \\
    & Szymanowicz et al. \cite{szymanowicz2022discrete} & 89.2 & 88.3 & - \\
    & Wu et al. \cite{wu2022self} & - & - & \underline{80.5} \\
    & Leroux et al. \cite{leroux2022multi} & - & 88.3 & - \\
    & Deng et al. \cite{deng2023bi} & \underline{98.9} & \uwave{ 89.7 } & 75.0 \\
    &  Lv et al. \cite{lv2021learning} & 96.9 & 89.5 & 73.8 \\
    &  Szymanowicz et al. \cite{Szymanowicz2021XMANEM} & 84.4 & 75.3 & 70.4 \\    
    \midrule
    & {Ours} & {\bf 99.3} & {\bf 90.7} & {\bf 82.2} \\
    \bottomrule
     \end{tabular} 
    }
   \vspace{0.5cm}
    \quad
   \subfloat[]{%
  \begin{tabular}{clccc}
    \toprule
    & \textbf{Method (Number of Tasks)} & \textbf{UCSD} & \textbf{Avenue} & \textbf{SHTech} \\
    \midrule
     \multirow{12}{*}{\rotatebox[origin=c]{90}{Multi-task}} &  Nguyen et al. \cite{nguyen2019anomaly} (2) & 96.2 & 86.9 & -  \\
    &  Lai et al. \cite{lai2020video} (2) & 95.8 & 87.4 & - \\ 
   &  Ionescu et al. \cite{ionescu2019object} (3) & 94.3 & 87.4 & 78.7 \\
   &  Abati et al. \cite{abati2019latent} (2) & 95.4 & - & 72.5 \\
   &  Chang et al. \cite{chang2020clustering} (3) & 96.5 & 86.0 & 73.3  \\
   &  Yu et al. \cite{yu2020cloze}  (2) & 97.3 & 90.2 & 74.8 \\
 &  Georgescu et al. \cite{georgescu2021background} (3) & 98.7 & \uwave{ 92.3 } & \uwave{ 82.7 } \\
   &  Cai et al. \cite{cai2021appearance} (2) &  96.6 & 86.6 & 73.7 \\
   &  Georgescu et al. \cite{georgescu2021anomaly}  (4) & {\bf 99.8} & 91.5 & 82.4 \\
   &  Doshi et al. \cite{doshi2023towards}  (2) & - & 79.0 & 68.9 \\
   & Reiss et al. \cite{reiss2022attribute} (3)  &  \uwave{ 99.1 } & \underline{93.3}  & {\bf 85.9} \\
 &  Barbalau et al. \cite{BARBALAU2023103656} (9) & - & {\bf 93.7} & \underline{83.8}  \\
    \midrule
   &  {Ours} &  \underline{99.3} & 90.7 &  82.2 \\
    \bottomrule
  \end{tabular}
  }
\end{table}

{\bf 
Single-task Framework.} Table~\ref{tab:auc} compares the frame-level AUC values of our framework (when operating as a single-task framework) against several OCC-based SOTA methods on UCSD  \cite{mahadevan2010anomaly}, Avenue \cite{lu2013abnormal}, and SHTech \cite{luo2017revisit} datasets. To ensure a fair comparison, we categorize these SOTA methods into two parts: Table~\ref{tab:auc}(a) for single-task methods and Table~\ref{tab:auc}(b) for multi-task methods. This categorization is imperative since multi-task methods usually exhibit superior performance by aggregating multiple single-task methods, as discussed in Section~\ref{sec:intro}. 

In Table~\ref{tab:auc}(a), our framework consistently outperforms all existing single-task methods on each benchmark. Specifically, it exceeds the previous best results of Deng et al. \cite{deng2023bi} by 0.4\% AUC on UCSD, Samuel et al. \cite{samuel2021svd} by 0.9\% AUC on Avenue, and Wu et al. \cite{wu2022self} by 1.7\% AUC on SHTech. Like us, Deng et al. \cite{deng2023bi} also employ a hybrid framework with middle-frame prediction as the proxy task. Their method cascades two U-Nets and a memory module, where optical flow is estimated indirectly as an intermediate step. This cascading structure introduces additional loss terms, making hyperparameter tuning more laborious than in our framework. While Deng et al. \cite{deng2023bi} perform comparably to ours on smaller datasets like UCSD and Avenue, they distinctly underperform on the more comprehensive SHTech. This highlights our framework's superior robustness in diverse scenarios, including complex real-world scenes in SHTech. Moreover, our framework firmly surpasses two previous VAD transformers \cite{feng2021convolutional,yuan2021transanomaly} on all benchmarks without requiring extra GAN schemes or sophisticated alterations to the behaviors of the original transformer. In other words, our framework better unleashes the potential of using vision transformers in VAD.

In Table~\ref{tab:auc}(b), our framework distinguishes itself even when compared with multi-task methods. It stands as one of the few methods \cite{georgescu2021anomaly,reiss2022attribute} that exceed the thresholds of 99\% AUC on UCSD, 90\% AUC on Avenue, and 80\% AUC on SHTech. Especially, it ranks {\it second} on UCSD and {\it fifth} on both Avenue and SHTech among all multi-task methods. In other words, our framework outperforms more than half of the existing multi-task methods using only one proxy task. This achievement challenges the dominance of works from the research group of Ionescu and Georgescu et al. \cite{ionescu2019object,georgescu2021anomaly,georgescu2021background,BARBALAU2023103656} for the first time, whose frameworks rely on up to nine proxy tasks \cite{BARBALAU2023103656} and an extensive range of pre-trained external networks. In the following paragraphs, we provide a detailed comparison with three notable multi-task methods.


Georgescu et al.'s leading work, SSMTL \cite{georgescu2021anomaly}, involves four proxy tasks and integrates YOLOv3 \cite{redmon2018yolov3} and ResNet-50 \cite{he2016deep} for object-level detection, followed by the fusion of both object- and frame-level anomaly scores at the final stage. This design results in an expansive model with over $88$M parameters, whereas our framework consists of only $8.5$M parameters and focuses solely on an independent prediction task. In addition, when YOLOv3 is eliminated, SSMTL reports 92.4\% AUC on UCSD and 86.9\% AUC on Avenue \cite{georgescu2021anomaly}, falling behind our framework by 6.9\% AUC on UCSD and 3.8\% AUC on Avenue. 

Barbalau et al.'s SSMTL++ \cite{BARBALAU2023103656} further extends Georgescu et al.'s SSMTL\cite{georgescu2021anomaly} by incorporating five new tasks and multiple facilitating modules, such as SelFlow \cite{8953810} for motion detection, Mask-RCNN \cite{he2017mask} for segmentation, and UniPose \cite{9157597} for pose estimation. While these external modules undoubtedly enhance SSMTL++'s AUC performance, they also increase its scale and complexity, weakening its independence and adaptability. However, our single-task framework still delivers competitive AUC scores. Notably, on the most comprehensive SHTech dataset, our framework is only 1.6\% behind SSMTL++ (82.2\% versus 83.8\%). Considering that SSMTL++ employs {\it eight more} proxy tasks than ours, we believe our performance is commendable.

Similarly, Reiss et al.'s AI-VAD \cite{reiss2022attribute} also relies heavily on off-the-shelf modules, including FlowNet2.0 \cite{ilg2017flownet}, Mask-RCNN \cite{he2017mask}, AlphaPose \cite{8237518}, and CLIP encoder \cite{radford2021learning}. These modules specialize in different video attributes and are pre-trained on vast datasets for their corresponding proxy tasks. Despite these efforts, our framework achieves comparable performance to AI-VAD, even surpassing it on UCSD (99.3\% versus 99.1\%). 

Table~\ref{tab:streetauc} compares the frame-level AUC value of our framework against
several OCC-based SOTA methods on StreetScene  \cite{ramachandra2020street}. As one of the largest and most intricate datasets,    
StreetScene has deterred many previous works from attempting evaluations. Even among the limited number of works \cite{lu2013abnormal,hasan2016learning,ramachandra2020street,cruz2022examination} that have tackled it, their performances are generally less satisfying \cite{Ramachandra2020ASO, georgescu2021background}. Remarkably, our framework outperforms these works with a superior AUC score of 61.9\%, lifting the current SOTA by 0.9\% AUC.


Table \ref{tab:rbdc} compares the TBDC/RBDC scores of our framework against existing methods on the UCSD  \cite{mahadevan2010anomaly}, Avenue \cite{lu2013abnormal}, and SHTech \cite{luo2017revisit} datasets. To calculate TBDC/RBDC, we identify detected regions by placing bounding boxes around pixels with large prediction errors through thresholding and connectivity analysis. This approach differs from other works \cite{ionescu2019object,georgescu2021anomaly,georgescu2021background,BARBALAU2023103656} that use object detectors like YOLOv3 \cite{redmon2018yolov3} to generate candidate regions, as our framework does not depend on any external detectors.  In addition,  we set $\alpha = 0.1$ and  $\beta = 0.1$ in TBDC/RBDC \cite{ramachandra2020street}, aligning with established practices in this field to ensure fair comparisons.


In general, Table \ref{tab:rbdc} reveals that the TBDC/RBDC performance of our single-task framework is on par with the multi-task frameworks proposed by the research group of Ionescu and Georgescu et al. \cite{ionescu2019object,georgescu2021anomaly,georgescu2021background,BARBALAU2023103656} across all benchmarks. In terms of TBDC, our framework surpasses the SoTA achieved by Georgescu et al. \cite{georgescu2021background} by 1.4\% on UCSD. It also ranks
second on Avenue with 67.3\% and third on SHTech with 79.1\%. Considering that our framework employs fewer proxy tasks and relies on no external modules compared to its competitors, we believe these achievements are quite notable. On the other hand, our RBDC performance, while not as numerically competitive as our TBDC, still secures second place on UCSD with 64.7\% and third place on Avenue with 48.1\%. We attribute this to the absence of object detectors in our framework, which may lead to lower overlaps between detected regions and the ground truth. Overall, the TBDC/RBDC performance of our single-task framework has reached the current SoTA levels achieved by multi-task frameworks, proving its competitiveness in localizing abnormal regions across frames. 

While we report TBDC/RBDC performance, these scores have inherent limitations. First, TBDC/RBDC are seldom implemented in the VAD field, with most reports coming from the research group of Ionescu and Georgescu et al. \cite{ionescu2019object,georgescu2021anomaly,georgescu2021background,BARBALAU2023103656}, as shown in Table \ref{tab:rbdc}, potentially leading to biased comparisons. Second, TBDC/RBDC can be adjusted to align with the frame-level criterion through simple post-processing techniques \cite{doshi2023towards,doshi2022modular}, and the data related to their implementation are not officially released \cite{Park2022FastAnoFA}. Third, the commonly-used parameter settings of $\alpha = 0.1$ and  $\beta = 0.1$ for TBDC/RBDC are quite lenient. Specifically, a region candidate is deemed an accurate detection under these metrics, as long as it overlaps more than {\it 10\%} of the ground truth in track or region. This lenient criterion may be impractical or less meaningful in real-world applications.


\begin{table}

  \caption{Frame-level AUC (\%) comparison with State-of-the-art OCC-based methods on StreetScene  \cite{ramachandra2020street}. 
  }
  \label{tab:streetauc}
  \centering
    \setlength{\tabcolsep}{4pt}
  \begin{tabular}{lc}
    \toprule
    \textbf{Method} & \textbf{StreetScene} \\
    \midrule
    Lu et al. \cite{lu2013abnormal} & \hspace{1pt} 48.0$^*$ \\
    Hasan et al. \cite{hasan2016learning} & \hspace{1pt} 61.0$^*$ \\
    Ramachandra et al. (Optical Flow) \cite{ramachandra2020street} &  51.0 \\
    Ramachandra et al. (Foreground Mask) \cite{ramachandra2020street} & 61.0 \\
     Cruz-Esquivel et al. \cite{cruz2022examination} & 57.3 \\
    \midrule
    Ours & 61.9 \\
    \bottomrule
    \multicolumn{2}{l}{\tiny * These results are reproduced by Ramachandra et al.\cite{Ramachandra2020ASO} on StreetScene.}

  \end{tabular}
\end{table}

\begin{table}

  \caption{RBDC/TBDC (\%) comparison with State-of-the-art OCC-based methods on UCSD  \cite{mahadevan2010anomaly}, Avenue \cite{lu2013abnormal}, and SHTech \cite{luo2017revisit}. The \textbf{best}, \underline{second-best}, and \uwave{third-best} performances are highlighted.}
  \label{tab:rbdc}
  \centering
    \setlength{\tabcolsep}{1.7pt}
  \begin{tabular}{clcccccc}
    \toprule
    & \multirow{2}{*}{\textbf{Method (Number of Tasks)}} & \multicolumn{2}{c}{\textbf{UCSD}} & \multicolumn{2}{c}{\textbf{Avenue}} & \multicolumn{2}{c}{\textbf{SHTech}} \\
    \cmidrule(lr){3-4} \cmidrule(lr){5-6} \cmidrule(lr){7-8}
     &  & RBDC & TBDC & RBDC & TBDC & RBDC & TBDC  \\
    \midrule
    &  Ionescu et al. \cite{ionescu2019object} (3) & \uwave{ 52.8 } & \uwave{ 72.9 } & 15.8 & 27.0 & 20.7 & 44.5 \\
 &  Georgescu et al. \cite{georgescu2021background} (3) & {\bf 69.2} & \underline{93.2} & {\bf 65.1} & \uwave{ 66.9 } & \uwave{ 41.3 } & 78.8 \\
       &  Georgescu et al. \cite{georgescu2021anomaly}  (4) & - & - & \underline{57.0} & 58.3 & \underline{42.8} & \underline{83.9} \\
        & Barbalau et al. \cite{BARBALAU2023103656} (9) & - & - & 40.9 & {\bf 82.1} & {\bf 47.1} & {\bf 85.6} \\
    \midrule
 &  Ours & \underline{64.7} & {\bf 94.6} & \uwave{ 48.1 } & \underline{67.3} & 38.9 & \uwave{ 79.1 } \\
    \bottomrule
  \end{tabular}
\end{table}

{\bf Multi-task Branch.} Table~\ref{tab:georgescucomp} validates that our framework can serve as an effective branch in multi-task approaches. To this end, we leverage Georgescu et al.'s SSMTL \cite{georgescu2021anomaly} as the baseline due to its current peak performance and abundant experimental results for reference. We first rigorously examine their results on UCSD and Avenue by implementing the code from their official repositories. Since their experiments did not include SHTech, we meticulously execute their code on SHTech and present the results as values marked with an asterisk ($*$) in Table~\ref{tab:georgescucomp}. 
Subsequently, we replace their middle-frame prediction branch with our own framework. Considering the inherent architectural disparities, we devise two integration mechanisms as illustrated in Table~\ref{tab:georgescucomp}: (1) ${\text{\it Ours}}_{1}$ represents the object-level adaptation of our proposed framework. It incorporates YOLOv3 as an object detector and aligns in scale with the other three object-level branches from \cite{georgescu2021anomaly}. In this case, all four branches share the same spatial encoder, followed by a 3D CNN which remains shared among branches from \cite{georgescu2021anomaly}. (2) ${\text{\it Ours}}_{2}$ represents the proposed framework, which generates predictions at the frame level. It remains structurally independent, while being trained jointly with the other branches from \cite{georgescu2021anomaly}. Please note that we follow the ablation study conducted by \cite{georgescu2021anomaly} to present the object-level performance of their four-task framework in Table~\ref{tab:georgescucomp}, differing from their fused object-and-frame-level performance shown in Table~\ref{tab:auc}(b).

The results in Table~\ref{tab:georgescucomp} confirm the effectiveness of our single-task framework in optimizing multi-task methods based on the following observations. First, our frame-level framework (${\text{\it Ours}}_{2}$) significantly outperforms all individual branches in SSMTL \cite{georgescu2021anomaly}, particularly their corresponding middle-frame prediction task (branch {\it C}) by 2.2\% AUC on UCSD, 7.2\% AUC on Avenue, and 14.5\% on SHTech. Second, our object-level variant (${\text{\it Ours}}_{1}$) generally performs slightly worse than ${\text{\it Ours}}_{2}$ due to YOLOv3's missed detections  \cite{georgescu2021background}, but it still solidly defeats any branch from \cite{georgescu2021anomaly}, surpassing their branch {\it C} by 0.3\% AUC on UCSD, 4.8\% AUC on Avenue, and 14.2\% on SHTech. Third, when integrated with various branches, both ${\text{\it Ours}}_{1}$ and ${\text{\it Ours}}_{2}$ bring consistent enhancements across all different task combinations. Interestingly, while the object-level ${\text{\it Ours}}_{1}$ generally underperforms the frame-level ${\text{\it Ours}}_{2}$, the difference gap is less pronounced on SHTech compared to UCSD or Avenue. ${\text{\it Ours}}_{1}$ even marginally surpasses ${\text{\it Ours}}_{2}$ when paired solely with the arrow of time task (branch {\it A}), achieving AUC scores of 83.6\% versus 83.5\% on SHTech. We believe this is due to the high-resolution nature of the SHTech dataset, which alleviates the impact of YOLOv3’s missed detections and sometimes aids in locating subtle anomalies. 
Moreover, increasing the number of tasks from 3 to 4 yields only marginal improvements on UCSD, indicating a potential performance saturation for this specific dataset.

\begin{table}
   \caption{Frame-level AUC (\%) on UCSD \cite{mahadevan2010anomaly}, Avenue \cite{lu2013abnormal}, and SHTech \cite{luo2017revisit} obtained by replacing the middle-frame prediction branch in SSMTL \cite{georgescu2021anomaly} with our proposed framework. The best performances are in \textbf{bold}.
  }
  \label{tab:georgescucomp}
  \centering
  \setlength{\tabcolsep}{4pt}
  \begin{tabular}{cccccccccc}
    
\toprule
\multirow{2}{1.08cm}{\bf Number of Tasks} & \multicolumn{6}{c}{\bf Proxy Task Branch} & \multicolumn{3}{c}{\bf AUC} \\
\cmidrule(lr){2-7} \cmidrule(l){8-10} 
& ${\text{\it A}}$ & ${\text{\it B}}$  & ${\text{\it C}}$  & ${\text{\it D}}$  & ${\text{\it Ours}}_{1}$ & ${\text{\it Ours}}_{2}$ & UCSD & Avenue & SHTech \\
\midrule
1& \twemoji{check mark} &{\bf -}&{\bf -}&{\bf -}&{\bf -}&{\bf -}& 89.4 & 83.6 & 69.3$^{*}$ \\
1&{\bf -}&\twemoji{check mark} &{\bf -}&{\bf -}&{\bf -}&{\bf -}& 94.9 & 83.4 & 72.8$^{*}$ \\
1&{\bf -}&{\bf -}&\twemoji{check mark}&{\bf -}&{\bf -}&{\bf -}& 97.1 & 83.5 & 67.7$^{*}$ \\
1&{\bf -}&{\bf -}&{\bf -}&\twemoji{check mark}&{\bf -}&{\bf -}& 97.1 & 73.7 & 63.2$^{*}$ \\
1&{\bf -}&{\bf -}&{\bf -}&{\bf -}&\twemoji{check mark}&{\bf -}& 97.4& 88.3 & 81.9\phantom{$^{*}$}\\
1&{\bf -}&{\bf -}&{\bf -}&{\bf -}&{\bf -}&\twemoji{check mark}& {\bf 99.3} & {\bf 90.7} & {\bf 82.2}\phantom{$^{*}$} \\
\midrule
2&\twemoji{check mark}&{\bf -}&\twemoji{check mark}&{\bf -}&{\bf -}&{\bf -}& 97.7& 84.2 & 73.5$^{*}$ \\
2&\twemoji{check mark}&{\bf -}&{\bf -}&{\bf -}&\twemoji{check mark}&{\bf -}& 98.5 & 89.1 & {\bf 83.6}\phantom{$^{*}$} \\
2&\twemoji{check mark}&{\bf -}&{\bf -}&{\bf -}&{\bf -}&\twemoji{check mark}& {\bf 99.6} & {\bf 91.4} & 83.5\phantom{$^{*}$} \\
\midrule
3&\twemoji{check mark}&\twemoji{check mark}&\twemoji{check mark}&{\bf -}&{\bf -}&{\bf -}& 98.8 & 90.7 & 75.9$^{*}$ \\
3&\twemoji{check mark}&\twemoji{check mark}&{\bf -}&{\bf -}&\twemoji{check mark}&{\bf -}& 99.3& 91.5 & 83.8\phantom{$^{*}$}\\
3&\twemoji{check mark}&\twemoji{check mark}&{\bf -}&{\bf -}&{\bf -}&\twemoji{check mark}& {\bf 99.8} & {\bf 92.0} & {\bf 84.1}\phantom{$^{*}$} \\
\midrule
4&\twemoji{check mark}&\twemoji{check mark}&\twemoji{check mark}&\twemoji{check mark}&{\bf -}&{\bf -}& 99.8 & 91.9 & 76.1$^{*}$ \\
4&\twemoji{check mark}&\twemoji{check mark}&{\bf -}&\twemoji{check mark}&\twemoji{check mark}&{\bf -}& 99.6 & 92.0 & 84.0\phantom{$^{*}$}   \\
4&\twemoji{check mark}&\twemoji{check mark}&{\bf -}&\twemoji{check mark}&{\bf -}&\twemoji{check mark}& {\bf 99.8} & {\bf 93.3} & {\bf 84.6}\phantom{$^{*}$} \\

\bottomrule
\multicolumn{10}{l}{\tiny Note that {\it A}, {\it B},  {\it C}, and {\it D} refer to different branches of object-level tasks introduced by Georgescu et al. \cite{georgescu2021anomaly}, } \\[-0.5em]
\multicolumn{10}{l}{\tiny namely, the arrow of time, motion irregularity, middle-frame prediction, and model distillation. ${\text{\it Ours}}_{1}$ denotes} \\[-0.5em]
\multicolumn{10}{l}{\tiny the object-level variant of our framework, which incorporates YOLOv3, while ${\text{\it Ours}}_{2}$ denotes our proposed} \\[-0.5em]
\multicolumn{10}{l}{\tiny frame-level framework that omits YOLOv3.}

  \end{tabular}
\end{table}


\subsection{Pixel-level Localization}

\begin{figure*}[t]
\centering
\includegraphics[width=\textwidth]{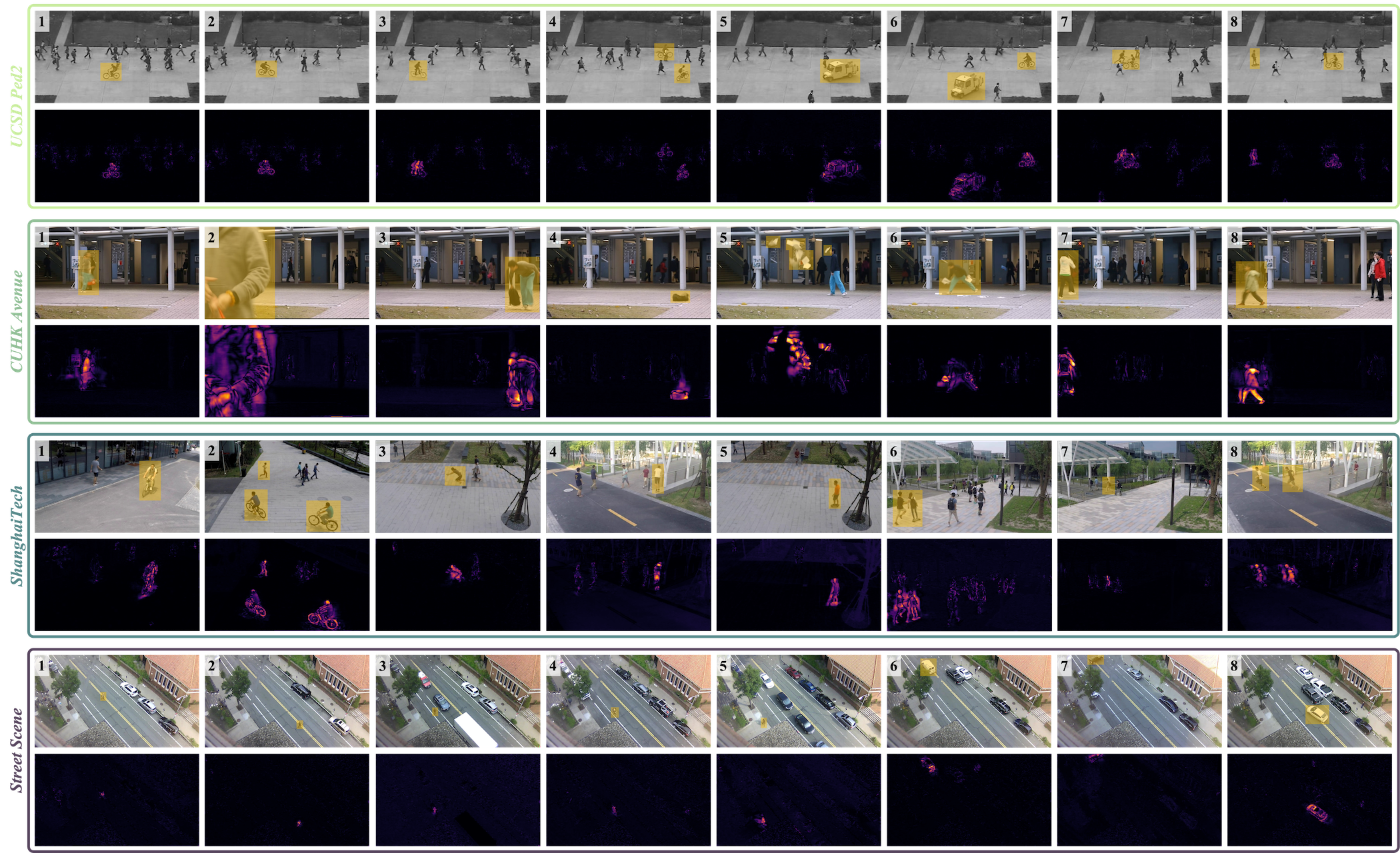}
\caption{Pixel-level anomaly detection instances on UCSD  \cite{mahadevan2010anomaly}, Avenue \cite{lu2013abnormal}, SHTech \cite{luo2017revisit}, and StreetScene \cite{ramachandra2020street} datasets, presented from top to bottom. For each dataset, the first row displays the ground-truth frames, with ground-truth anomalies highlighted in yellow boxes. The second row exhibits the prediction error maps generated by our framework, where brighter colors indicate larger prediction errors.} 
\label{fig:pixellevel}
\end{figure*}

Figure~\ref{fig:pixellevel} presents a set of anomaly detection instances from our benchmark tests, visualized as error maps. 
These instances offer a clear insight into our framework's ability to discriminate and locate anomalies at the pixel level, demonstrating its robustness across a wide range of anomaly types. From this figure, we can see that our framework adeptly handles straightforward anomalies, which aligns with other established works, including: (1) diverse anomalous objects like the vehicle in UCSD sample 5, the falling backpack in Avenue sample 4, and flying papers in Avenue sample 5; (2) unusual activities, such as cycling in UCSD sample 3, running in Avenue sample 1, dancing in Avenue sample 7, jumping in SHTech sample 3, and skateboarding in SHTech sample 5 and StreetScene sample 3.

More importantly, our framework excels in addressing intricate scenarios that are currently challenging many existing works: (1) crowded scenes with multiple anomalies and complex dynamics, as seen in SHTech samples 2 and 6; (2) anomalies in varying scales, exemplified by the size difference between running individuals in SHTech samples 7 and 8; (3) anomalies observed from various viewpoints,  illustrated by cycling events captured from different camera angles in SHTech samples 1 and 2; (4) context-dependent anomalies, such as pedestrians jaywalking in StreetScene samples 1 and 2, bicycles deviating from designated lanes in  StreetScene samples 4 and 5, and vehicles crossing solid lines in StreetScene samples 6 and 8.

\subsection{Ablation Study}
\label{subsec:dicuss}


{\bf Influence of Positional Encoding.}  Positional encoding, often deemed essential in most transformer-based frameworks, is rigorously examined in this study. To this end, we integrate trigonometric position functions \cite{vaswani2017attention} into both our uni- and bi-directional frameworks using two plausible configurations: (1) {\it channel-wise} summation with feature map $\mathcal{E}_{i}$, which involves 64 sinusoids, and (2) {\it element-wise} summation with flattened $\mathcal{E}_{i}$, which involves $64\times 64\times 64$ sinusoids. All experiments are performed on the UCSD and Avenue datasets. Note that we differentiate the framework's directionality in the experiments as it affects the continuity of context clips. The results, as summarized in Table~\ref{tab:pe}, reveal that both positional encoding schemes gain AUC performance similar to the baseline. This observation suggests that injecting positional knowledge holds negligible significance to our framework. In other words, our framework can infer positional knowledge directly from the feature map $\mathcal{E}_{i}$. One possible explanation is that video clips always exhibit gradual variations during the evolution of frames, unlike word tokens in NLP tasks, which are often arranged repetitively and randomly. Moreover,  our video clips are also considerably shorter in length compared to sentences with up to 512 word tokens, thereby restraining the occurrence of complex positional relationships.  

The  Model Fit Level (MFL) metric measures the optimal validation error when early stopping occurs, serving as an indicator of the framework's prediction capability. Table~\ref{tab:pe} shows that the MFL of each configuration  is generally proportional to their respective AUC performance. This indicates that middle-frame prediction closely aligns with the target VAD task, affirming its suitability as a proxy task. This assertion finds further support in the subsequent experiments.

\begin{table}
   \caption{AUC (\%) and MFL ($\times10^{-3}$) of different positional encoding schemes on UCSD \cite{mahadevan2010anomaly} and Avenue \cite{lu2013abnormal}.}
  \label{tab:pe}
  \centering
  \setlength{\tabcolsep}{3.5pt}
  \begin{tabular}{cccccccc}
\toprule
\multicolumn{2}{c}{\bf Directionality}&\multicolumn{2}{c}{\bf Positional Encoding} & \multicolumn{2}{c}{\bf UCSD} &\multicolumn{2}{c}{\bf Avenue} \\
\cmidrule(r){1-2} \cmidrule(lr){3-4} \cmidrule(l){5-8}
 Uni. & Bi. & Channel-wise & Element-wise & AUC & MFL  & AUC& MFL\\
\midrule
 \twemoji{check mark} &  {\bf -} &  {\bf -} &  {\bf -} & 97.3 & 10.5  & 88.1& 17.3\\
  \twemoji{check mark} &  {\bf -} & \twemoji{check mark}&   {\bf -} & 97.0 & 11.0  & 88.4 & 15.6\\
  \twemoji{check mark} &  {\bf -} & {\bf -} & \twemoji{check mark} & 96.8 & 9.8  & 87.9 & 17.9\\
    {\bf -} & \twemoji{check mark} & {\bf -} &  {\bf -} & 99.3 & 9.2  & 90.7 & 12.9\\
    {\bf -} & \twemoji{check mark} & \twemoji{check mark}&  {\bf -} & 98.8 & 8.9  & 89.6 & 12.1\\
    {\bf -} & \twemoji{check mark} & {\bf -}  & \twemoji{check mark} & 98.2 & 9.1  & 90.3 & 13.3\\
 
\bottomrule
\multicolumn{8}{l}{\tiny   Uni. = Uni-directional. Bi. = Bi-directional.}

  \end{tabular}
\end{table}

\begin{figure}[t]
  \centering
  \includegraphics[width=1.01\linewidth]{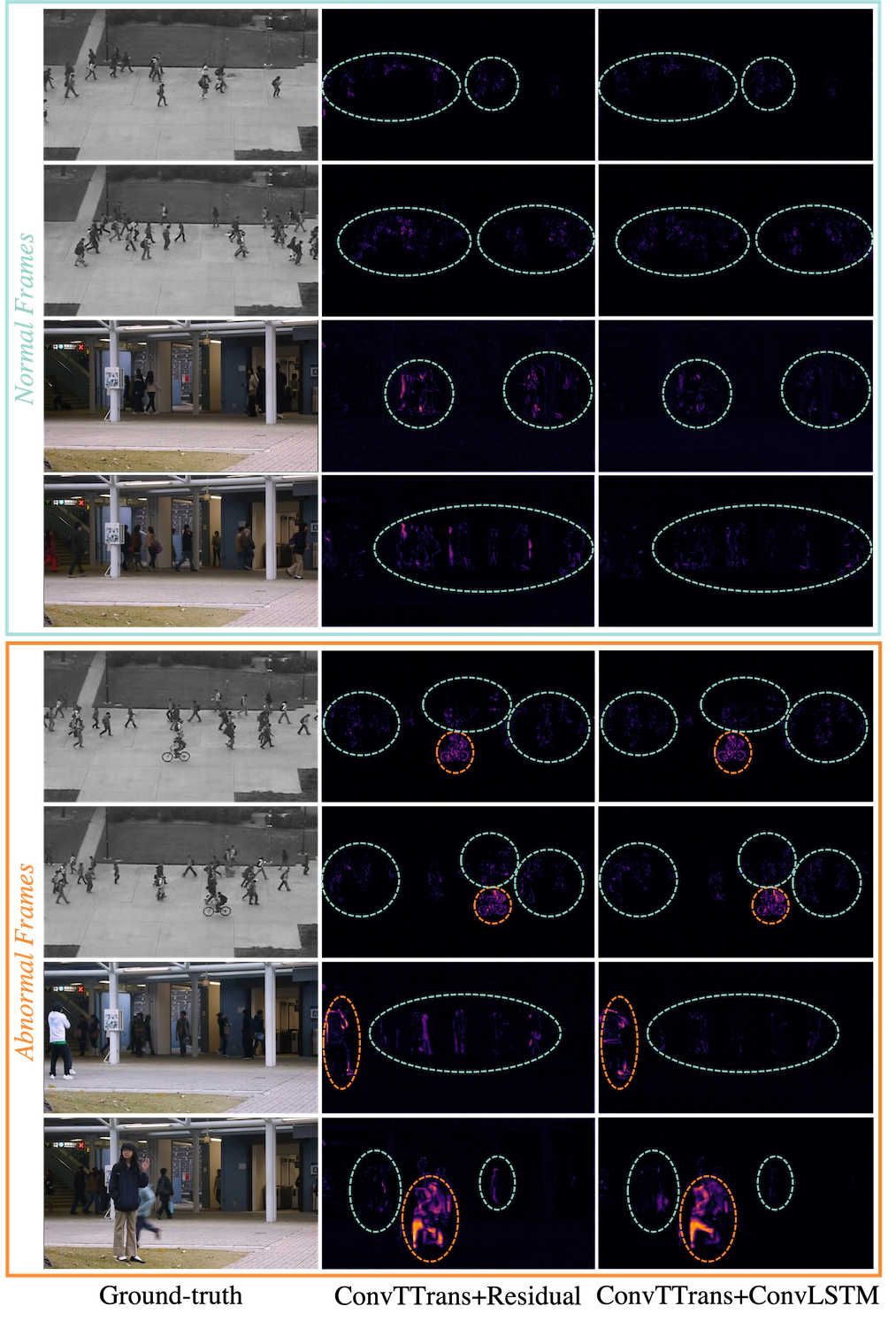}
   \caption{LI-ConvLSTM bridge versus residual connection on UCSD \cite{mahadevan2010anomaly} and Avenue \cite{lu2013abnormal}. From left to right, we show the ground-truth frames, prediction error maps generated by ConvTTrans with residual connection, and prediction error maps generated by ConvTTrans with LI-ConvLSTM bridge. Normal events are circled in green, while abnormal events are circled in orange.}  
   \label{fig:lowlevel}
\end{figure}

{\bf Impact of Low-level Features.}  This study seeks to evaluate the efficacy of our LI-ConvLSTM bridge and the importance of utilizing low-level features. To this end, we establish two comparative setups: one excludes the LI-ConvLSTM bridge from our framework, while the other replaces it with the classic residual connection. For simplicity, all  configurations in our experiments adhere to the uni-directional framework. The results reported in Table~\ref{tab:lowlevel}  showcase that incorporating the LI-ConvLSTM bridge elevates the  baseline (ConvTTrans-only) by 1.6\% AUC on UCSD and 2.1\% AUC on Avenue. This manifests the significance of low-level features on the final AUC performance in our MAE+SSIM-based inference system. In addition, the LI-ConvLSTM bridge outperforms the residual connection by 0.5\% AUC on UCSD and 0.9\% AUC on Avenue, even though  both methods skip the same number of intermediate layers. 
 To explain this observation, Fig.~\ref{fig:lowlevel} reveals that when compared to the residual connection (middle), the LI-ConvLSTM bridge (right) is more capable of reducing prediction errors in normal regions while maintaining those in abnormal regions. Considering their architectural differences, we infer that the LI-ConvLSTM bridge can better leverage the (shallow) spatio-temporal dynamics in low-level features to enrich predictions with finer details.

\begin{table}
  \caption{AUC (\%) and MFL ($\times10^{-3}$) of different low-level feature transfer mechanisms on UCSD \cite{mahadevan2010anomaly} and Avenue \cite{lu2013abnormal}.}
  \label{tab:lowlevel}
  \centering
  \setlength{\tabcolsep}{4pt}
  \begin{tabular}{ccccccc}
\toprule
\multicolumn{3}{c}{\bf Module}& \multicolumn{2}{c}{\bf UCSD} &\multicolumn{2}{c}{\bf Avenue} \\
\cmidrule(r){1-3} \cmidrule(lr){4-5} \cmidrule(l){6-7}
 ConvTTrans & Residual & ConvLSTM & AUC & MFL  & AUC& MFL\\
\midrule
 \twemoji{check mark} &   {\bf -} &   {\bf -} & 95.7 & 23.5  & 86.0 & 25.9 \\
   \twemoji{check mark} &  \twemoji{check mark} &   {\bf -} & 96.8 & 13.2  & 87.2 & 18.2 \\
      \twemoji{check mark} &   {\bf -} &  \twemoji{check mark} & 97.3 & 10.5  & 88.1 & 17.3 \\
  
\bottomrule

  \end{tabular}
\end{table}

\begin{figure}[t]
  \hspace{-6pt}
 \includegraphics[width=1.0\linewidth]{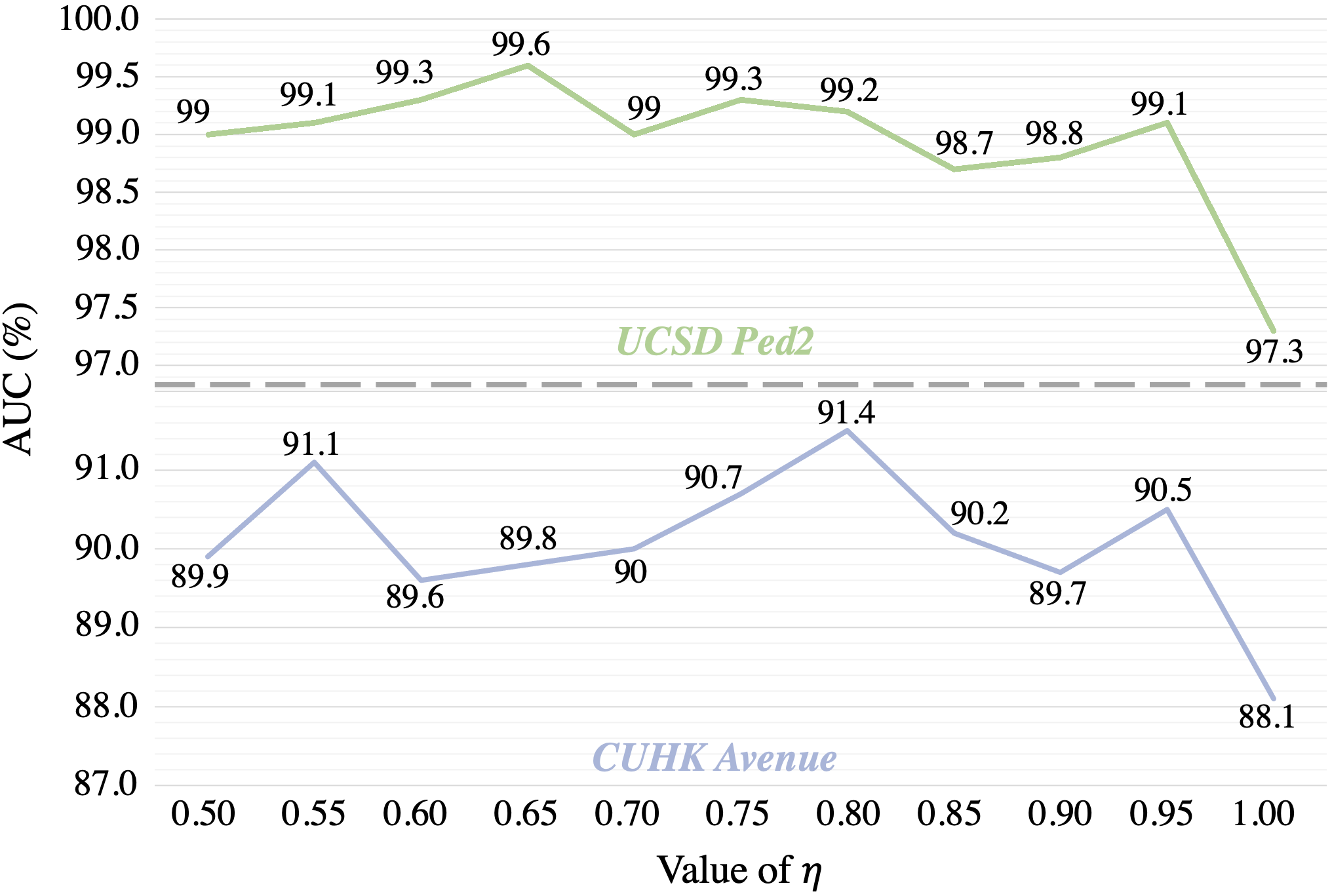}
   \caption{Frame-level AUC performance with respect to varying weighting factor $\eta$  on UCSD \cite{mahadevan2010anomaly} and Avenue \cite{lu2013abnormal}.}
   \label{fig:biweight}
\end{figure}

{\bf Bi-directional Prediction.} 
This study aims to investigate the influence of the bi-directional mechanism and its overall benefits to the VAD task compared to uni-directional frameworks. To this end, we first examine three strategies for fusing predictions from both directions: concatenation, frame-wise weighted sum, and convolutional gated (Conv-gated) sum. The results in Table~\ref{tab:bidirect} illustrate that the frame-wise sum slightly outperforms the other two fusion strategies in terms of AUC and MFL, indicating that the forward and backward predictions conform to independent element-to-element correspondence. Based on this insight, we further fine-tune the weighting factor $\eta$ to gauge the relative significance of two prediction processes. As depicted in Fig.~\ref{fig:biweight}, all equispaced samples of $\eta$ perform on par with each other, while sharply surpassing the uni-directional baseline (where $\eta=1.00$). We thus reason that these two prediction processes can counterbalance each other through intermediate representations to reach a common optimum. Therefore, we {\it do not} prioritize the optimality of a specific $\eta$ value in our experiments. By reporting non-optimal results, we aim to derive more generalizable conclusions. In this work, our reported AUC performances, including those on UCSD ($ 99.3 \%$) and Avenue ($ 90.7 \%$), are all based on $\eta = 0.75$, which is sufficient but not optimal for any dataset.

With the above fine-tuned bi-directional framework, we thoroughly compare it against forward-only and backward-only frameworks across multiple datasets, and extend our experiments to include the more challenging SHTech and StreetScene datasets. The results 
in Table \ref{tab:unibi} reveal three key observations. First, both forward-only and backward-only frameworks cannot outperform the bi-directional framework in any of the benchmarks, highlighting the advantages of bi-directional prediction. Second, the backward-only framework generally achieves performance comparable to that of the forward-only framework, attesting to their symmetry in architecture and behavior. Third, upon closer examination, the backward-only framework {\it slightly} underperforms the forward-only framework. We believe this discrepancy is due to the fact that motion irregularities in the benchmarks, as well as in the real world, intrinsically convey the forward progression of time, which forward networks are inherently better equipped to reveal. 

\begin{figure}[t]

  \centering
  \includegraphics[width=\linewidth]{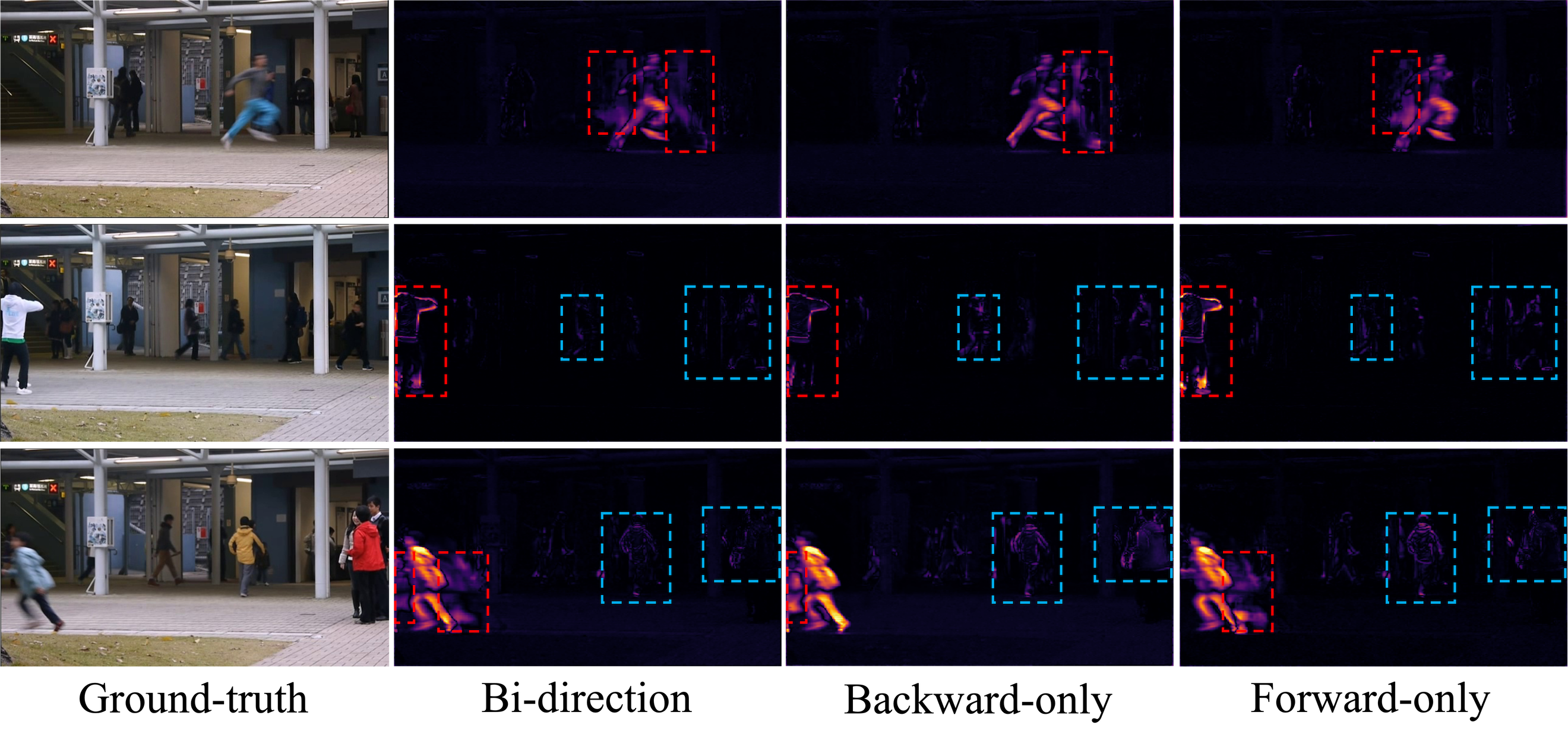}
  \caption{Illustrative examples of predictions from our bi-directional framework and its uni-directional variants on Avenue \cite{lu2013abnormal}. Normal regions are enclosed in blue rectangles, whereas abnormal regions are highlighted with red rectangles.}
\label{fig:unibiexamples}
\end{figure}

\begin{figure}[t]
  \hspace{-6pt}
 \includegraphics[width=1.0\linewidth]{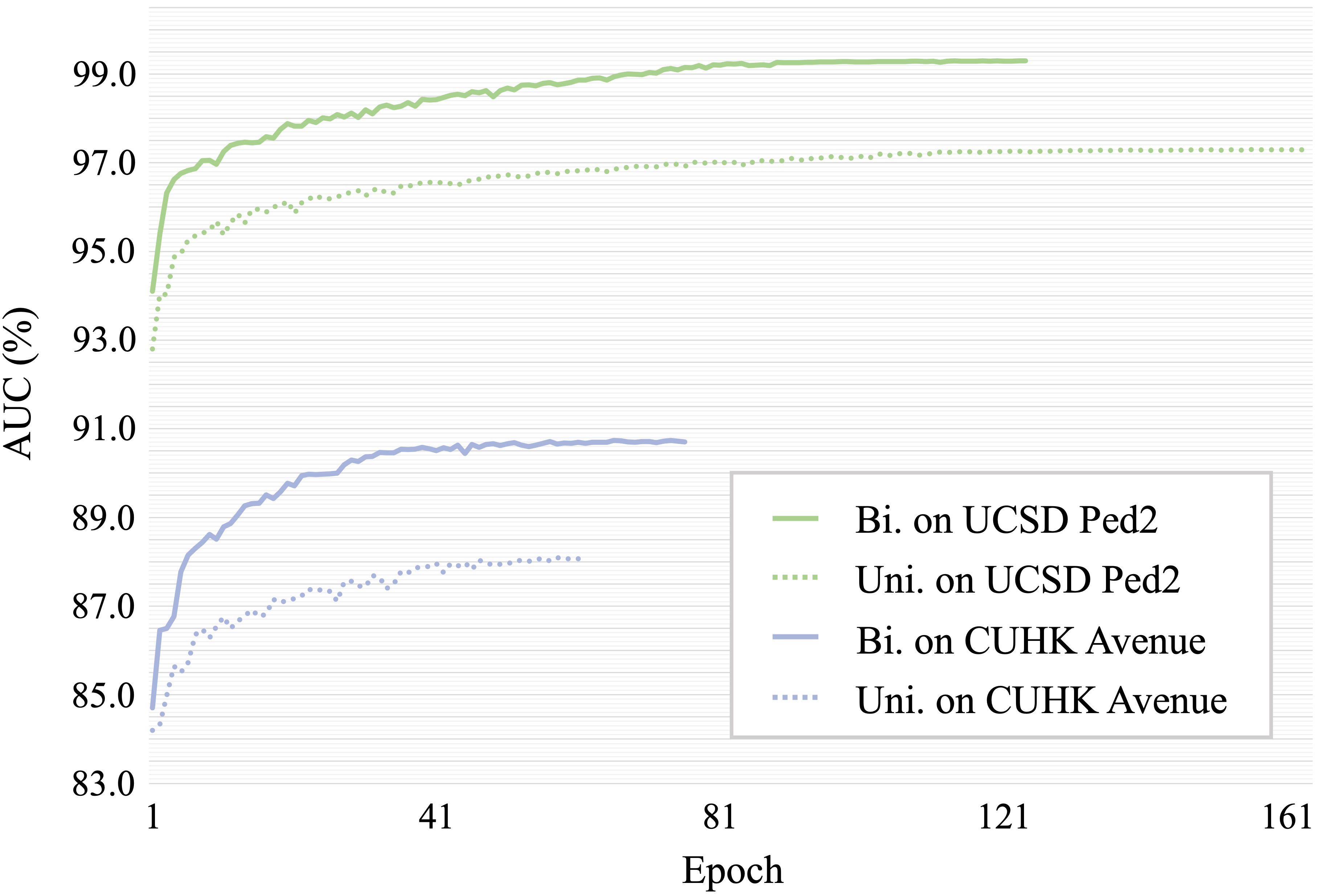}
   \caption{Convergence curves of frame-level AUC for uni- and bi-directional frameworks on UCSD \cite{mahadevan2010anomaly} and Avenue \cite{lu2013abnormal}. Note that $\eta$ is set to 0.75 in bi-directional frameworks. }
   \label{fig:converg}
\end{figure}

Prediction examples illustrated in Fig. \ref{fig:unibiexamples} further confirm the benefits of employing bi-directional frameworks over uni-directional ones. Specifically, the bi-directional framework efficiently suppresses prediction errors for normal events (blue regions) while maintaining those for abnormal events (red regions). We infer that prediction errors for normal events are relatively random in either forward or backward predictions. Therefore, combining predictions from both directions can partially cancel out these errors. In contrast, prediction errors for abnormal events are denser and more concentrated, so combining the two directions likely preserves or even intensifies these errors. Notably, for abnormal events with heavy motion irregularities (e.g., the running people on the top and bottom rows of Fig. \ref{fig:unibiexamples}), we can clearly observe that the bi-directional framework merges the separate motion trajectories predicted by the backward-only and forward-only frameworks, thereby amplifying the prediction errors and improving the effectiveness of anomaly detection. Moreover, Fig.~\ref{fig:converg} showcases that the increased temporal flexibility introduced by the bi-directional mechanism can considerably accelerate convergence during training.

\begin{table}
  \caption{AUC (\%) and MFL ($\times10^{-3}$) of different approaches to fusing bi-directional predictions on UCSD \cite{mahadevan2010anomaly} and Avenue \cite{lu2013abnormal}.}
  \label{tab:bidirect}
  \centering
  \setlength{\tabcolsep}{4pt}
  \begin{tabular}{ccccc}
\toprule
\multirow{2}{*}{\bf Fusion Mode}& \multicolumn{2}{c}{\bf UCSD} &\multicolumn{2}{c}{\bf Avenue} \\
\cmidrule(lr){2-3} \cmidrule(l){4-5}
   & AUC & MFL  & AUC& MFL\\
\midrule
Concatenation& 98.5 & 9.4 & 90.2 & 12.8 \\
Conv-gated Sum& 98.1 & 9.9 & 89.0 & 14.2 \\
Frame-wise Sum ($\eta=0.75$) & 99.3 & 9.2 & 90.7 & 12.9 \\
\bottomrule

  \end{tabular}
\end{table}

\begin{table}[htbp]

\caption{Frame-level AUC (\%) comparison of uni- and bi-directional frameworks.}
  \label{tab:unibi}
  \centering
  \begin{tabular}{lcccc}
    \toprule
   \textbf{Directionality} & \textbf{UCSD} & \textbf{Avenue} & \textbf{SHTech} & \textbf{StreetScene}\\
    \midrule
     Backward-only & 96.8 & 88.4 & 80.7 & 59.2\\
     Forward-only & 97.3 & 88.1 & 80.9 & 60.4 \\
     Bi-direction ($\eta$ = 0.75) &  99.3 & 90.7 & 82.2 & 61.9\\
      \bottomrule
     \end{tabular} 
    \end{table}

 \begin{figure}[t]
  \hspace{-6pt}
 \includegraphics[width=1.01\linewidth]{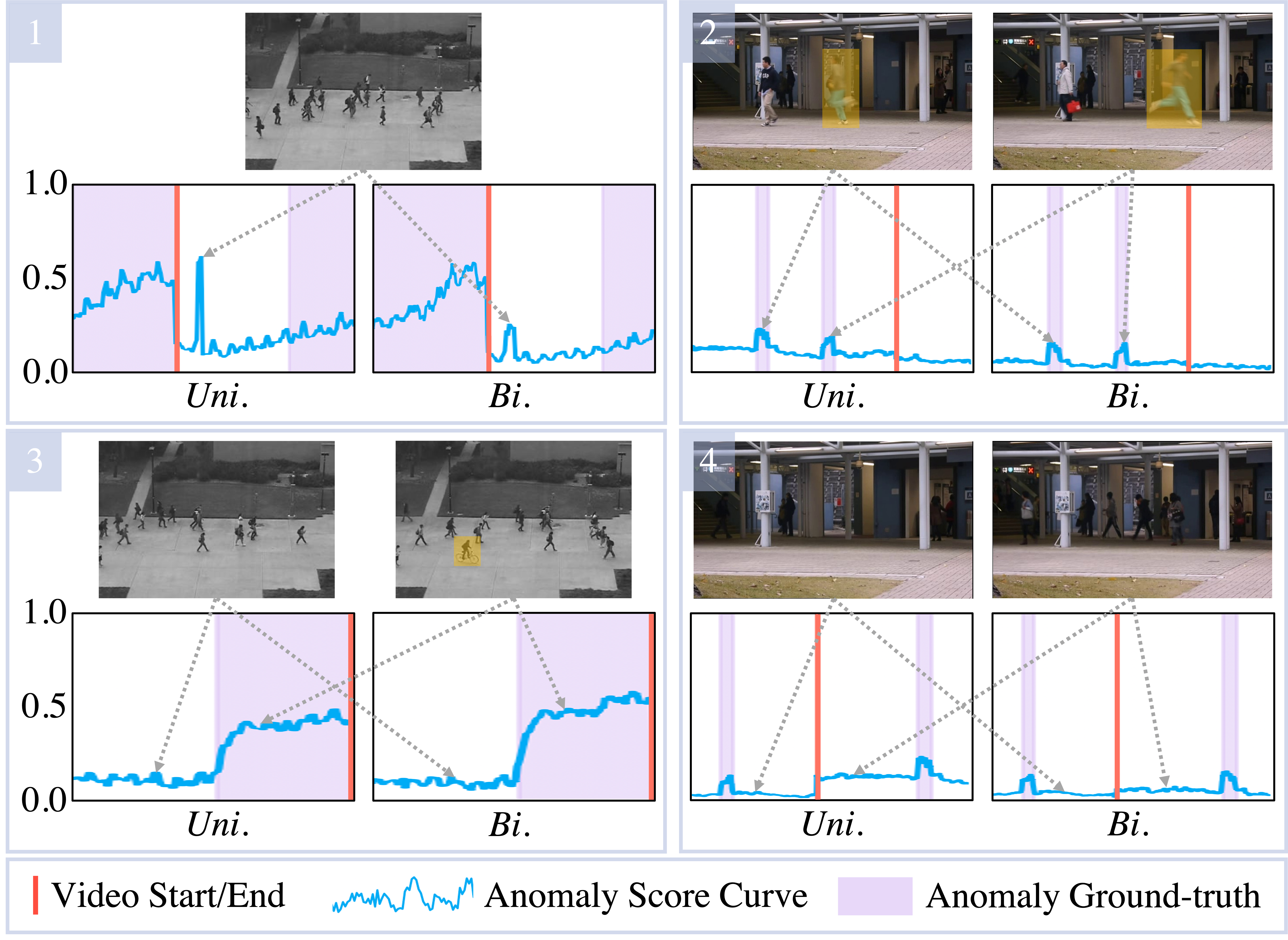}
   \caption{Examples of frame-level anomaly scores generated by the uni- and bi-directional frameworks on UCSD \cite{mahadevan2010anomaly} and Avenue \cite{lu2013abnormal}. 
   }
   \label{fig:bivisual}
\end{figure}

{\bf False Alarms and Stability.} This study exhibits the effectiveness of the bi-directional mechanism in mitigating false alarms and enhancing the stability of anomaly inference. To this end, we compare the anomaly scores generated by both uni-directional and bi-directional frameworks, and illustrate several examples across four distinct scenarios in Fig.~\ref{fig:bivisual}. In scenario 1 (upper-left), the bi-directional mechanism successfully alleviates a false alarm induced by motion disruption from a missing frame in UCSD. In scenario 2 (upper-right), the mechanism ensures consistent detection for recurring anomaly types on Avenue, exemplified by an individual running in opposite directions. In scenario 3 (lower-left), the mechanism widens the detection gap between normal and abnormal events on UCSD, despite the limited coverage of abnormal events. In scenario 4 (lower-right), the mechanism facilitates smoother detection during a sudden transition from sparse to dense scenes on Avenue, even though the latter contains more intricate dynamics and is inherently harder to predict. Overall, these scenarios disclose the bi-directional framework's confidence and robustness in anomaly inference, attributed to its utilization of backward temporal information.


 \begin{figure}[t]

  \hspace{-12pt}
 \includegraphics[width=1.05\linewidth]{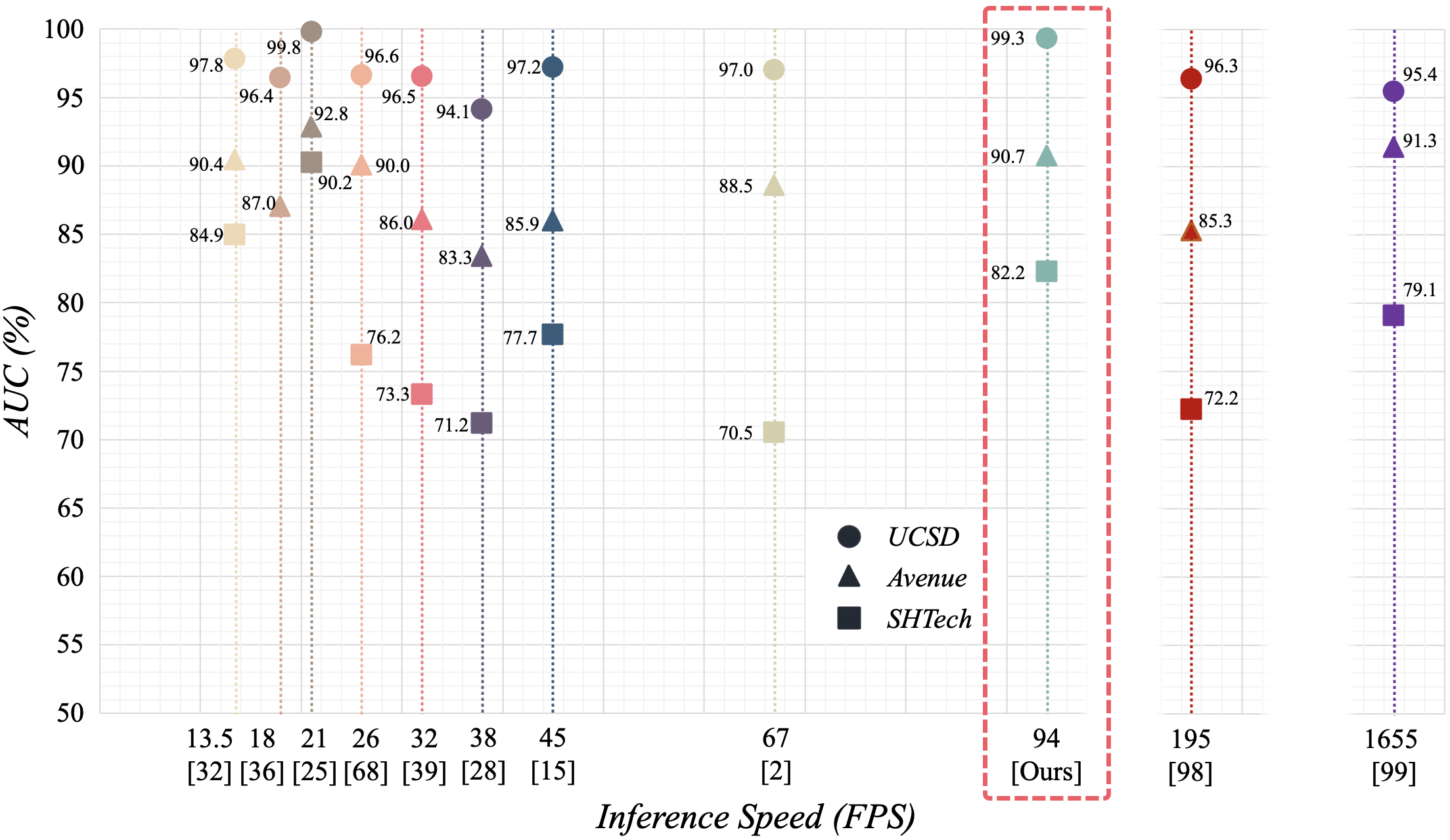}
   \caption{Comparison of frame-level AUC performance and inference speed on UCSD \cite{mahadevan2010anomaly}, Avenue \cite{lu2013abnormal}, and SHTech \cite{luo2017revisit}.}
   \label{fig:runtime}
\end{figure}

{\bf Complexity and Runtime.} The implementation of the convolutional transformer and LI-ConvLSTM bridge results in a limited scale of learnable parameters ($\approx8.5$M). When executed on an RTX A5000 GPU, our framework exhibits a rapid and tractable convergence process during training, eliminating the need for the pre-training techniques commonly used in other transformer-based approaches \cite{wang2022bevt}. During inference, our framework efficiently calculates anomaly scores with an average processing time of 10.37 ms (96 FPS) for a $256 \times 256$ grayscale frame and 10.65 ms (94 FPS) for a $256 \times 256$ RGB frame. As shown in Fig.~\ref{fig:runtime}, our inference speed of 94 FPS far surpasses the majority of previous methods, including the speed of 13.5 FPS in \cite{ionescu2019object},  18 FPS in \cite{yuan2021transanomaly}, 21 FPS in \cite{georgescu2021anomaly}, 26 FPS in \cite{lee2019bman}, 32 FPS in \cite{chang2020clustering}, 38 FPS in \cite{gong2019memorizing}, 45 FPS in \cite{feng2021convolutional}, and 67 FPS in \cite{park2020learning}. This remarkable speed is achieved while assuring our framework's capacity to deliver comparable or superior AUC performance. 

One may observe from Fig.~\ref{fig:runtime} that our inference speed of 94 FPS is slower than the 195 FPS reported by FastAno \cite{9706649} and the 1655 FPS by Self-Distilled MAE\cite{ristea2024self}. This discrepancy can be explained by three aspects: (1) Both FastAno and Self-Distilled MAE achieve lower AUC performance compared to ours, as shown in Fig.~\ref{fig:runtime}. They prioritize high inference speeds as their primary innovation but sacrifice a degree of detection accuracy for speed. Specifically, they accelerate the inference process by eliminating certain modules \cite{9706649}, using fewer blocks than the original prototypes \cite{ristea2024self}, or freezing specific parameters \cite{ristea2024self}. (2) Both FastAno and Self-Distilled MAE are tested on an NVIDIA GeForce RTX 3090. The RTX 3090, which has a memory bandwidth of 936.2 GB/s, is approximately
20\% faster than our A5000, which has a bandwidth of 768.0 GB/s. (3)  Despite their high experimental speeds, the read-in speed limit of $30 \sim 60$ FPS in most standard surveillance systems implies that our framework has similar inference speeds to FastAno and Self-Distilled MAE in real-world applications. 

While our framework exhibits remarkable speed, we should note that 
its bi-directional mechanism endures a constant latency of 12 frames from $\mathcal{X}_{t+1}$ to $\mathcal{X}_{t+12}$. If we consider 30 FPS as the read-in speed of mainstream surveillance cameras, the inference lapse would be  0.40 seconds, around 1.60 times the average human response of 0.25 seconds.  One feasible solution to this is to remove the entire backward decoding pipeline at the expense of introducing more fluctuations in the anomaly scores. Alternatively, one could use cameras equipped with higher frame rates, such as 60 FPS, to reduce the lapse proportionally to 0.20 seconds.

\section{Conclusion}
\label{sec:conclusion}

This paper introduced a bi-directional hybrid framework for anomaly detection in videos, with middle-frame prediction as the primary proxy task. Specifically, our framework adopts a bi-directional structure, which processes frames in both forward and backward directions, to fully exploit the temporal domain and thus enhance detection stability. In addition, our framework leverages a new convolutional transformer to effectively model the temporal dependencies among frames and generate frame-wise predictions. Furthermore, it incorporates a layer-interactive ConvLSTM bridge to enable the propagation of low-level features across different layers and time-steps, thereby improving the prediction accuracy for fine details.  After being trained end-to-end as a one-class classifier with no ancillaries,  our framework consistently delivered competing performance across various public benchmarks, whether employed as a standalone single-task approach or integrated as a branch in a multi-task approach. Therefore, this work not only provides an alternative to existing VAD approaches but also demonstrates the potential to advance their performance by optimizing frameworks for individual proxy tasks. Moreover, it highlights the strengths of combining vision transformers and ConvLSTMs in the VAD context.

\bibliographystyle{IEEEtran}
\bibliography{main}

\begin{thebibliography}{10}
\providecommand{\url}[1]{#1}
\csname url@samestyle\endcsname
\providecommand{\newblock}{\relax}
\providecommand{\bibinfo}[2]{#2}
\providecommand{\BIBentrySTDinterwordspacing}{\spaceskip=0pt\relax}
\providecommand{\BIBentryALTinterwordstretchfactor}{4}
\providecommand{\BIBentryALTinterwordspacing}{\spaceskip=\fontdimen2\font plus
\BIBentryALTinterwordstretchfactor\fontdimen3\font minus \fontdimen4\font\relax}
\providecommand{\BIBforeignlanguage}[2]{{%
\expandafter\ifx\csname l@#1\endcsname\relax
\typeout{** WARNING: IEEEtran.bst: No hyphenation pattern has been}%
\typeout{** loaded for the language `#1'. Using the pattern for}%
\typeout{** the default language instead.}%
\else
\language=\csname l@#1\endcsname
\fi
#2}}
\providecommand{\BIBdecl}{\relax}
\BIBdecl

\bibitem{zaheer2020old}
M.~Z. Zaheer, J.-h. Lee, M.~Astrid, and S.-I. Lee, ``Old is gold: Redefining the adversarially learned one-class classifier training paradigm,'' in \emph{IEEE/CVF Conference on Computer Vision and Pattern Recognition}, 2020, pp. 14\,183--14\,193.

\bibitem{park2020learning}
H.~Park, J.~Noh, and B.~Ham, ``Learning memory-guided normality for anomaly detection,'' in \emph{IEEE/CVF Conference on Computer Vision and Pattern Recognition}, 2020, pp. 14\,372--14\,381.

\bibitem{feng2021mist}
J.-C. Feng, F.-T. Hong, and W.-S. Zheng, ``Mist: Multiple instance self-training framework for video anomaly detection,'' in \emph{IEEE/CVF Conference on Computer Vision and Pattern Recognition}, 2021, pp. 14\,009--14\,018.

\bibitem{zhong2019graph}
J.-X. Zhong, N.~Li, W.~Kong, S.~Liu, T.~H. Li, and G.~Li, ``Graph convolutional label noise cleaner: Train a plug-and-play action classifier for anomaly detection,'' in \emph{IEEE/CVF Conference on Computer Vision and Pattern Recognition}, 2019, pp. 1237--1246.

\bibitem{wan2020weakly}
B.~Wan, Y.~Fang, X.~Xia, and J.~Mei, ``Weakly supervised video anomaly detection via center-guided discriminative learning,'' in \emph{IEEE International Conference on Multimedia and Expo (ICME)}, 2020, pp. 1--6.

\bibitem{sultani2018real}
W.~Sultani, C.~Chen, and M.~Shah, ``Real-world anomaly detection in surveillance videos,'' in \emph{IEEE/CVF Conference on Computer Vision and Pattern Recognition}, 2018, pp. 6479--6488.

\bibitem{zhang2019temporal}
J.~Zhang, L.~Qing, and J.~Miao, ``Temporal convolutional network with complementary inner bag loss for weakly supervised anomaly detection,'' in \emph{IEEE International Conference on Image Processing (ICIP)}, 2019, pp. 4030--4034.

\bibitem{zaheer2020claws}
M.~Z. Zaheer, A.~Mahmood, M.~Astrid, and S.-I. Lee, ``Claws: Clustering assisted weakly supervised learning with normalcy suppression for anomalous event detection,'' in \emph{IEEE European Conference on Computer Vision}, 2020, pp. 358--376.

\bibitem{zaheer2022generative}
M.~Z. Zaheer, A.~Mahmood, M.~H. Khan, M.~Segu, F.~Yu, and S.-I. Lee, ``Generative cooperative learning for unsupervised video anomaly detection,'' in \emph{IEEE/CVF Conference on Computer Vision and Pattern Recognition}, 2022, pp. 14\,744--14\,754.

\bibitem{pang2020self}
G.~Pang, C.~Yan, C.~Shen, A.~v.~d. Hengel, and X.~Bai, ``Self-trained deep ordinal regression for end-to-end video anomaly detection,'' in \emph{IEEE/CVF Conference on Computer Vision and Pattern Recognition}, 2020, pp. 12\,173--12\,182.

\bibitem{Lin_Chen_Li_Yu_2022}
X.~Lin, Y.~Chen, G.~Li, and Y.~Yu, ``A causal inference look at unsupervised video anomaly detection,'' in \emph{AAAI conference on artificial intelligence}, 2022, pp. 1620--1629.

\bibitem{tudor2017unmasking}
R.~Tudor~Ionescu, S.~Smeureanu, B.~Alexe, and M.~Popescu, ``Unmasking the abnormal events in video,'' in \emph{IEEE/CVF International Conference on Computer Vision}, 2017, pp. 2895--2903.

\bibitem{wang2018detecting}
S.~Wang, Y.~Zeng, Q.~Liu, C.~Zhu, E.~Zhu, and J.~Yin, ``Detecting abnormality without knowing normality: A two-stage approach for unsupervised video abnormal event detection,'' in \emph{ACM International Conference on Multimedia}, 2018, pp. 636--644.

\bibitem{doshi2020continual}
K.~Doshi and Y.~Yilmaz, ``Continual learning for anomaly detection in surveillance videos,'' in \emph{IEEE/CVF Conference on Computer Vision and Pattern Recognition Workshops}, 2020, pp. 1025--1034.

\bibitem{feng2021convolutional}
X.~Feng, D.~Song, Y.~Chen, Z.~Chen, J.~Ni, and H.~Chen, ``Convolutional transformer based dual discriminator generative adversarial networks for video anomaly detection,'' in \emph{ACM International Conference on Multimedia}, 2021, pp. 5546--5554.

\bibitem{liu2019margin}
W.~Liu, W.~Luo, Z.~Li, P.~Zhao, and S.~Gao, ``Margin learning embedded prediction for video anomaly detection with a few anomalies,'' in \emph{International Joint Conference on Artificial Intelligence}, 2019, pp. 3023--3030.

\bibitem{zhu2022towards}
Y.~Zhu, W.~Bao, and Q.~Yu, ``Towards open set video anomaly detection,'' in \emph{IEEE European Conference on Computer Vision}, 2022, p. 395–412.

\bibitem{deng2023bi}
H.~Deng, Z.~Zhang, S.~Zou, and X.~Li, ``Bi-directional frame interpolation for unsupervised video anomaly detection,'' in \emph{IEEE/CVF Winter Conference on Applications of Computer Vision}, 2023, pp. 2634--2643.

\bibitem{Szymanowicz2021XMANEM}
S.~Szymanowicz, J.~Charles, and R.~Cipolla, ``X-man: Explaining multiple sources of anomalies in video,'' in \emph{IEEE/CVF Conference on Computer Vision and Pattern Recognition Workshops}, 2021, pp. 3218--3226.

\bibitem{leroux2022multi}
S.~Leroux, B.~Li, and P.~Simoens, ``Multi-branch neural networks for video anomaly detection in adverse lighting and weather conditions,'' in \emph{IEEE/CVF Winter Conference on Applications of Computer Vision}, 2022, pp. 3027--3035.

\bibitem{szymanowicz2022discrete}
S.~Szymanowicz, J.~Charles, and R.~Cipolla, ``Discrete neural representations for explainable anomaly detection,'' in \emph{IEEE/CVF Winter Conference on Applications of Computer Vision}, 2022, pp. 1506--1514.

\bibitem{Park2022FastAnoFA}
C.~Park, M.~Cho, M.~Lee, and S.~Lee, ``Fastano: Fast anomaly detection via spatio-temporal patch transformation,'' in \emph{IEEE/CVF Winter Conference on Applications of Computer Vision}, 2022, pp. 1908--1918.

\bibitem{lv2021learning}
H.~Lv, C.~Chen, Z.~Cui, C.~Xu, Y.~Li, and J.~Yang, ``Learning normal dynamics in videos with meta prototype network,'' in \emph{IEEE/CVF Conference on Computer Vision and Pattern Recognition}, 2021, pp. 15\,425--15\,434.

\bibitem{cai2021appearance}
R.~Cai, H.~Zhang, W.~Liu, S.~Gao, and Z.~Hao, ``Appearance-motion memory consistency network for video anomaly detection,'' in \emph{AAAI conference on artificial intelligence}, 2021, pp. 938--946.

\bibitem{georgescu2021anomaly}
M.-I. Georgescu, A.~Barbalau, R.~T. Ionescu, F.~S. Khan, M.~Popescu, and M.~Shah, ``Anomaly detection in video via self-supervised and multi-task learning,'' in \emph{IEEE/CVF Conference on Computer Vision and Pattern Recognition}, 2021, pp. 12\,742--12\,752.

\bibitem{liu2018future}
W.~Liu, W.~Luo, D.~Lian, and S.~Gao, ``Future frame prediction for anomaly detection--a new baseline,'' in \emph{IEEE/CVF Conference on Computer Vision and Pattern Recognition}, 2018, pp. 6536--6545.

\bibitem{tang2020integrating}
Y.~Tang, L.~Zhao, S.~Zhang, C.~Gong, G.~Li, and J.~Yang, ``Integrating prediction and reconstruction for anomaly detection,'' \emph{Pattern Recognition Letters}, vol. 129, pp. 123--130, 2020.

\bibitem{gong2019memorizing}
D.~Gong, L.~Liu, V.~Le, B.~Saha, M.~R. Mansour, S.~Venkatesh, and A.~v.~d. Hengel, ``Memorizing normality to detect anomaly: Memory-augmented deep autoencoder for unsupervised anomaly detection,'' in \emph{IEEE/CVF International Conference on Computer Vision}, 2019, pp. 1705--1714.

\bibitem{nguyen2019anomaly}
T.-N. Nguyen and J.~Meunier, ``Anomaly detection in video sequence with appearance-motion correspondence,'' in \emph{IEEE/CVF International Conference on Computer Vision}, 2019, pp. 1273--1283.

\bibitem{ouyang2021video}
Y.~Ouyang and V.~Sanchez, ``Video anomaly detection by estimating likelihood of representations,'' in \emph{International Conference on Pattern Recognition (ICPR)}, 2021, pp. 8984--8991.

\bibitem{Xu2015LearningDR}
D.~Xu, E.~Ricci, Y.~Yan, J.~Song, and N.~Sebe, ``Learning deep representations of appearance and motion for anomalous event detection,'' in \emph{British Machine Vision Conference (BMVC)}, 2015, pp. 8.1--8.12.

\bibitem{ionescu2019object}
R.~T. Ionescu, F.~S. Khan, M.-I. Georgescu, and L.~Shao, ``Object-centric auto-encoders and dummy anomalies for abnormal event detection in video,'' in \emph{IEEE/CVF Conference on Computer Vision and Pattern Recognition}, 2019, pp. 7842--7851.

\bibitem{shen2022video}
G.~Shen, Y.~Ouyang, and V.~Sanchez, ``Video anomaly detection via prediction network with enhanced spatio-temporal memory exchange,'' in \emph{IEEE International Conference on Acoustics, Speech and Signal Processing (ICASSP)}, 2022, pp. 3728--3732.

\bibitem{liu2021hybrid}
Z.~Liu, Y.~Nie, C.~Long, Q.~Zhang, and G.~Li, ``A hybrid video anomaly detection framework via memory-augmented flow reconstruction and flow-guided frame prediction,'' in \emph{IEEE/CVF International Conference on Computer Vision}, 2021, pp. 13\,588--13\,597.

\bibitem{lai2020video}
Y.~Lai, R.~Liu, and Y.~Han, ``Video anomaly detection via predictive autoencoder with gradient-based attention,'' in \emph{IEEE International Conference on Multimedia and Expo (ICME)}, 2020, pp. 1--6.

\bibitem{yuan2021transanomaly}
H.~Yuan, Z.~Cai, H.~Zhou, Y.~Wang, and X.~Chen, ``Transanomaly: Video anomaly detection using video vision transformer,'' \emph{IEEE Access}, vol.~9, pp. 123\,977--123\,986, 2021.

\bibitem{luo2017remembering}
W.~Luo, W.~Liu, and S.~Gao, ``Remembering history with convolutional lstm for anomaly detection,'' in \emph{IEEE International Conference on Multimedia and Expo (ICME)}, 2017, pp. 439--444.

\bibitem{georgescu2021background}
M.~I. Georgescu, R.~T. Ionescu, F.~S. Khan, M.~Popescu, and M.~Shah, ``A background-agnostic framework with adversarial training for abnormal event detection in video,'' \emph{IEEE Transactions on Pattern Analysis and Machine Intelligence}, vol.~44, no.~9, pp. 4505--4523, 2021.

\bibitem{chang2020clustering}
Y.~Chang, Z.~Tu, W.~Xie, and J.~Yuan, ``Clustering driven deep autoencoder for video anomaly detection,'' in \emph{IEEE European Conference on Computer Vision}, 2020, pp. 329--345.

\bibitem{abati2019latent}
D.~Abati, A.~Porrello, S.~Calderara, and R.~Cucchiara, ``Latent space autoregression for novelty detection,'' in \emph{IEEE/CVF Conference on Computer Vision and Pattern Recognition}, 2019, pp. 481--490.

\bibitem{yu2020cloze}
G.~Yu, S.~Wang, Z.~Cai, E.~Zhu, C.~Xu, J.~Yin, and M.~Kloft, ``Cloze test helps: Effective video anomaly detection via learning to complete video events,'' in \emph{ACM International Conference on Multimedia}, 2020, p. 583–591.

\bibitem{doshi2023towards}
K.~Doshi and Y.~Yilmaz, ``Towards interpretable video anomaly detection,'' in \emph{IEEE/CVF Winter Conference on Applications of Computer Vision}, 2023, pp. 2655--2664.

\bibitem{Caruana1997MultitaskL}
R.~Caruana, ``Multitask learning,'' \emph{Machine Learning}, vol.~28, pp. 41--75, 1997.

\bibitem{dosovitskiy2020image}
A.~Dosovitskiy, L.~Beyer, A.~Kolesnikov, D.~Weissenborn, X.~Zhai, T.~Unterthiner, M.~Dehghani, M.~Minderer, G.~Heigold, S.~Gelly \emph{et~al.}, ``An image is worth 16x16 words: Transformers for image recognition at scale,'' in \emph{International Conference on Learning Representations}, 2021.

\bibitem{arnab2021vivit}
A.~Arnab, M.~Dehghani, G.~Heigold, C.~Sun, M.~Lu{\v{c}}i{\'c}, and C.~Schmid, ``Vivit: A video vision transformer,'' in \emph{IEEE/CVF International Conference on Computer Vision}, 2021, pp. 6836--6846.

\bibitem{wang2022bevt}
R.~Wang, D.~Chen, Z.~Wu, Y.~Chen, X.~Dai, M.~Liu, Y.-G. Jiang, L.~Zhou, and L.~Yuan, ``Bevt: Bert pretraining of video transformers,'' in \emph{IEEE/CVF Conference on Computer Vision and Pattern Recognition}, 2022, pp. 14\,733--14\,743.

\bibitem{mahadevan2010anomaly}
V.~Mahadevan, W.~Li, V.~Bhalodia, and N.~Vasconcelos, ``Anomaly detection in crowded scenes,'' in \emph{IEEE/CVF Conference on Computer Vision and Pattern Recognition}, 2010, pp. 1975--1981.

\bibitem{lu2013abnormal}
C.~Lu, J.~Shi, and J.~Jia, ``Abnormal event detection at 150 fps in matlab,'' in \emph{IEEE/CVF International Conference on Computer Vision}, 2013, pp. 2720--2727.

\bibitem{luo2017revisit}
W.~Luo, W.~Liu, and S.~Gao, ``A revisit of sparse coding based anomaly detection in stacked rnn framework,'' in \emph{IEEE/CVF International Conference on Computer Vision}, 2017, pp. 341--349.

\bibitem{ramachandra2020street}
B.~Ramachandra and M.~Jones, ``Street scene: A new dataset and evaluation protocol for video anomaly detection,'' in \emph{IEEE/CVF Winter Conference on Applications of Computer Vision}, 2020, pp. 2558--2567.

\bibitem{cao2024advanced}
K.~Cao, T.~Zhang, and J.~Huang, ``Advanced hybrid lstm-transformer architecture for real-time multi-task prediction in engineering systems,'' \emph{Scientific Reports}, vol.~14, no.~1, p. 4890, 2024.

\bibitem{xu2021hybrid}
X.~Xu and X.~Zheng, ``Hybrid model for network anomaly detection with gradient boosting decision trees and tabtransformer,'' in \emph{IEEE International Conference on Acoustics, Speech and Signal Processing (ICASSP)}, 2021, pp. 8538--8542.

\bibitem{openai2023chatgpt}
OpenAI, ``Chatgpt: {GPT}-4 model,'' \url{https://chat.openai.com/}, 2023.

\bibitem{vaswani2017attention}
A.~Vaswani, N.~Shazeer, N.~Parmar, J.~Uszkoreit, L.~Jones, A.~N. Gomez, {\L}.~Kaiser, and I.~Polosukhin, ``Attention is all you need,'' in \emph{Advances in Neural Information Processing Systems}, 2017.

\bibitem{liu2021swin}
Z.~Liu, Y.~Lin, Y.~Cao, H.~Hu, Y.~Wei, Z.~Zhang, S.~Lin, and B.~Guo, ``Swin transformer: Hierarchical vision transformer using shifted windows,'' in \emph{IEEE/CVF International Conference on Computer Vision}, 2021, pp. 10\,012--10\,022.

\bibitem{fan2021multiscale}
H.~Fan, B.~Xiong, K.~Mangalam, Y.~Li, Z.~Yan, J.~Malik, and C.~Feichtenhofer, ``Multiscale vision transformers,'' in \emph{IEEE/CVF International Conference on Computer Vision}, 2021, pp. 6824--6835.

\bibitem{meng2022adavit}
L.~Meng, H.~Li, B.-C. Chen, S.~Lan, Z.~Wu, Y.-G. Jiang, and S.-N. Lim, ``Adavit: Adaptive vision transformers for efficient image recognition,'' in \emph{IEEE/CVF Conference on Computer Vision and Pattern Recognition}, 2022, pp. 12\,309--12\,318.

\bibitem{parmar2018image}
N.~Parmar, A.~Vaswani, J.~Uszkoreit, L.~Kaiser, N.~Shazeer, A.~Ku, and D.~Tran, ``Image transformer,'' in \emph{International Conference on Machine Learning}, 2018, pp. 4055--4064.

\bibitem{lee2022multi}
J.~Lee, W.-J. Nam, and S.-W. Lee, ``Multi-contextual predictions with vision transformer for video anomaly detection,'' in \emph{International Conference on Pattern Recognition (ICPR)}, 2022, pp. 1012--1018.

\bibitem{goodfellow2020generative}
I.~Goodfellow, J.~Pouget-Abadie, M.~Mirza, B.~Xu, D.~Warde-Farley, S.~Ozair, A.~Courville, and Y.~Bengio, ``Generative adversarial nets,'' in \emph{Advances in Neural Information Processing Systems}, vol.~27, 2014.

\bibitem{doshi2022modular}
K.~Doshi and Y.~Yilmaz, ``A modular and unified framework for detecting and localizing video anomalies,'' in \emph{IEEE/CVF Winter Conference on Applications of Computer Vision}, 2022, pp. 3982--3991.

\bibitem{wu2022self}
J.-C. Wu, H.-Y. Hsieh, D.-J. Chen, C.-S. Fuh, and T.-L. Liu, ``Self-supervised sparse representation for video anomaly detection,'' in \emph{IEEE European Conference on Computer Vision}, 2022, pp. 729--745.

\bibitem{samuel2021svd}
D.~J. Samuel and F.~Cuzzolin, ``Svd-gan for real-time unsupervised video anomaly detection,'' in \emph{British Machine Vision Conference (BMVC)}, 2021.

\bibitem{wang2020cluster}
Z.~Wang, Y.~Zou, and Z.~Zhang, ``Cluster attention contrast for video anomaly detection,'' in \emph{ACM International Conference on Multimedia}, 2020, p. 2463–2471.

\bibitem{shi2015convolutional}
X.~Shi, Z.~Chen, H.~Wang, D.-Y. Yeung, W.-K. Wong, and W.-c. Woo, ``Convolutional lstm network: A machine learning approach for precipitation nowcasting,'' in \emph{Advances in Neural Information Processing Systems}, vol.~28, 2015.

\bibitem{hochreiter1997long}
S.~Hochreiter and J.~Schmidhuber, ``Long short-term memory,'' \emph{Neural Computation}, vol.~9, no.~8, pp. 1735--1780, 1997.

\bibitem{wang2018abnormal}
L.~Wang, F.~Zhou, Z.~Li, W.~Zuo, and H.~Tan, ``Abnormal event detection in videos using hybrid spatio-temporal autoencoder,'' in \emph{IEEE International Conference on Image Processing (ICIP)}, 2018, pp. 2276--2280.

\bibitem{lee2019bman}
S.~Lee, H.~G. Kim, and Y.~M. Ro, ``Bman: bidirectional multi-scale aggregation networks for abnormal event detection,'' \emph{IEEE Transactions on Image Processing}, vol.~29, pp. 2395--2408, 2019.

\bibitem{song2019learning}
H.~Song, C.~Sun, X.~Wu, M.~Chen, and Y.~Jia, ``Learning normal patterns via adversarial attention-based autoencoder for abnormal event detection in videos,'' \emph{IEEE Transactions on Multimedia}, vol.~22, no.~8, pp. 2138--2148, 2019.

\bibitem{li2018videolstm}
Z.~Li, K.~Gavrilyuk, E.~Gavves, M.~Jain, and C.~G. Snoek, ``Videolstm convolves, attends and flows for action recognition,'' \emph{Computer Vision and Image Understanding}, vol. 166, pp. 41--50, 2018.

\bibitem{su2020convolutional}
J.~Su, W.~Byeon, J.~Kossaifi, F.~Huang, J.~Kautz, and A.~Anandkumar, ``Convolutional tensor-train lstm for spatio-temporal learning,'' in \emph{Advances in Neural Information Processing Systems}, vol.~33, 2020, pp. 13\,714--13\,726.

\bibitem{goodfellow2016deep}
I.~Goodfellow, Y.~Bengio, and A.~Courville, \emph{Deep learning}.\hskip 1em plus 0.5em minus 0.4em\relax MIT Press, 2016.

\bibitem{wang2017predrnn}
Y.~Wang, M.~Long, J.~Wang, Z.~Gao, and P.~S. Yu, ``Predrnn: Recurrent neural networks for predictive learning using spatiotemporal lstms,'' in \emph{Advances in Neural Information Processing Systems}, vol.~30, 2017.

\bibitem{wang2018predrnn++}
Y.~Wang, Z.~Gao, M.~Long, J.~Wang, and S.~Y. Philip, ``Predrnn++: Towards a resolution of the deep-in-time dilemma in spatiotemporal predictive learning,'' in \emph{International Conference on Machine Learning}, 2018, pp. 5123--5132.

\bibitem{8953605}
Y.~Wang, J.~Zhang, H.~Zhu, M.~Long, J.~Wang, and P.~S. Yu, ``Memory in memory: A predictive neural network for learning higher-order non-stationarity from spatiotemporal dynamics,'' in \emph{IEEE/CVF Conference on Computer Vision and Pattern Recognition}, 2019, pp. 9146--9154.

\bibitem{wang2004image}
Z.~Wang, A.~C. Bovik, H.~R. Sheikh, and E.~P. Simoncelli, ``Image quality assessment: from error visibility to structural similarity,'' \emph{IEEE Transactions on Image Processing}, vol.~13, no.~4, pp. 600--612, 2004.

\bibitem{Sgaard2016ApplicabilityOE}
J.~S{\o}gaard, L.~Krasula, M.~Shahid, D.~Temel, K.~Brunnstr{\"o}m, and M.~Razaak, ``Applicability of existing objective metrics of perceptual quality for adaptive video streaming,'' in \emph{Electronic Imaging}, 2016.

\bibitem{wang2011information}
Z.~Wang and Q.~Li, ``Information content weighting for perceptual image quality assessment,'' \emph{IEEE Transactions on Image Processing}, vol.~20, pp. 1185--1198, 2011.

\bibitem{wang2003multiscale}
Z.~Wang, E.~Simoncelli, and A.~Bovik, ``Multiscale structural similarity for image quality assessment,'' in \emph{Conference Record of the Asilomar Conference on Signals, Systems and Computers}, 2003, pp. 1398 -- 1402.

\bibitem{pytorch-ignite}
V.~Fomin, J.~Anmol, S.~Desroziers, J.~Kriss, and A.~Tejani, ``High-level library to help with training neural networks in pytorch,'' \url{https://github.com/pytorch/ignite}, 2020.

\bibitem{paszke2017automatic}
A.~Paszke, S.~Gross, S.~Chintala, G.~Chanan, E.~Yang, Z.~DeVito, Z.~Lin, A.~Desmaison, L.~Antiga, and A.~Lerer, ``Automatic differentiation in pytorch,'' 2017.

\bibitem{kingma2014adam}
D.~P. Kingma and J.~Ba, ``Adam: A method for stochastic optimization,'' in \emph{International Conference on Learning Representations}, 2015.

\bibitem{lu2020few}
Y.~Lu, F.~Yu, M.~K.~K. Reddy, and Y.~Wang, ``Few-shot scene-adaptive anomaly detection,'' in \emph{IEEE European Conference on Computer Vision}, 2020, pp. 125--141.

\bibitem{rodrigues2020multi}
R.~Rodrigues, N.~Bhargava, R.~Velmurugan, and S.~Chaudhuri, ``Multi-timescale trajectory prediction for abnormal human activity detection,'' in \emph{IEEE/CVF Winter Conference on Applications of Computer Vision}, 2020, pp. 2626--2634.

\bibitem{reiss2022attribute}
T.~Reiss and Y.~Hoshen, ``Attribute-based representations for accurate and interpretable video anomaly detection,'' \emph{arXiv preprint arXiv:2212.00789}, 2022.

\bibitem{BARBALAU2023103656}
A.~Barbalau, R.~T. Ionescu, M.~Georgescu, J.~Dueholm, B.~Ramachandra, K.~Nasrollahi, F.~Khan, T.~Moeslund, and M.~Shah, ``Ssmtl++: Revisiting self-supervised multi-task learning for video anomaly detection,'' \emph{Computer Vision and Image Understanding}, vol. 229, p. 103656, 2023.

\bibitem{redmon2018yolov3}
J.~Redmon and A.~Farhadi, ``Yolov3: An incremental improvement,'' \emph{arXiv preprint arXiv:1804.02767}, 2018.

\bibitem{he2016deep}
K.~He, X.~Zhang, S.~Ren, and J.~Sun, ``Deep residual learning for image recognition,'' in \emph{IEEE/CVF Conference on Computer Vision and Pattern Recognition}, 2016, pp. 770--778.

\bibitem{8953810}
P.~Liu, M.~Lyu, I.~King, and J.~Xu, ``Selflow: Self-supervised learning of optical flow,'' in \emph{IEEE/CVF Conference on Computer Vision and Pattern Recognition}, 2019, pp. 4566--4575.

\bibitem{he2017mask}
K.~He, G.~Gkioxari, P.~Doll{\'a}r, and R.~Girshick, ``Mask r-cnn,'' in \emph{IEEE/CVF International Conference on Computer Vision}, 2017, pp. 2961--2969.

\bibitem{9157597}
B.~Artacho and A.~Savakis, ``Unipose: Unified human pose estimation in single images and videos,'' in \emph{IEEE/CVF Conference on Computer Vision and Pattern Recognition}, 2020, pp. 7033--7042.

\bibitem{ilg2017flownet}
E.~Ilg, N.~Mayer, T.~Saikia, M.~Keuper, A.~Dosovitskiy, and T.~Brox, ``Flownet 2.0: Evolution of optical flow estimation with deep networks,'' in \emph{IEEE/CVF Conference on Computer Vision and Pattern Recognition}, 2017, pp. 2462--2470.

\bibitem{8237518}
H.-S. Fang, S.~Xie, Y.-W. Tai, and C.~Lu, ``Rmpe: Regional multi-person pose estimation,'' in \emph{IEEE International Conference on Computer Vision}, 2017, pp. 2353--2362.

\bibitem{radford2021learning}
A.~Radford, J.~W. Kim, C.~Hallacy, A.~Ramesh, G.~Goh, S.~Agarwal, G.~Sastry, A.~Askell, P.~Mishkin, J.~Clark \emph{et~al.}, ``Learning transferable visual models from natural language supervision,'' in \emph{International Conference on Machine Learning}, 2021, pp. 8748--8763.

\bibitem{hasan2016learning}
M.~Hasan, J.~Choi, J.~Neumann, A.~K. Roy-Chowdhury, and L.~S. Davis, ``Learning temporal regularity in video sequences,'' in \emph{IEEE/CVF Conference on Computer Vision and Pattern Recognition}, 2016, pp. 733--742.

\bibitem{cruz2022examination}
E.~Cruz-Esquivel and Z.~J. Guzman-Zavaleta, ``An examination on autoencoder designs for anomaly detection in video surveillance,'' \emph{IEEE Access}, vol.~10, pp. 6208--6217, 2022.

\bibitem{Ramachandra2020ASO}
B.~Ramachandra, M.~J. Jones, and R.~R. Vatsavai, ``A survey of single-scene video anomaly detection,'' \emph{IEEE Transactions on Pattern Analysis and Machine Intelligence}, vol.~44, pp. 2293--2312, 2020.

\bibitem{9706649}
C.~Park, M.~Cho, M.~Lee, and S.~Lee, ``Fastano: Fast anomaly detection via spatio-temporal patch transformation,'' in \emph{IEEE/CVF Winter Conference on Applications of Computer Vision}, 2022, pp. 1908--1918.

\bibitem{ristea2024self}
N.-C. Ristea, F.-A. Croitoru, R.~T. Ionescu, M.~Popescu, F.~S. Khan, M.~Shah \emph{et~al.}, ``Self-distilled masked auto-encoders are efficient video anomaly detectors,'' in \emph{IEEE/CVF Conference on Computer Vision and Pattern Recognition}, 2024, pp. 15\,984--15\,995.

\end{thebibliography}

\vfill

\end{document}